\documentclass[final]{cvpr}

\usepackage{times}
\usepackage{epsfig}
\usepackage{graphicx}
\usepackage{amsmath}
\usepackage{amssymb}

\usepackage{cuted}
\usepackage{capt-of}

\usepackage{color}
\usepackage{makecell}

\usepackage{overpic}
\usepackage{amssymb}
\usepackage{amsmath}
\usepackage{graphicx}

\usepackage{color}
\definecolor{turquoise}{cmyk}{0.65,0,0.1,0.3}
\definecolor{purple}{rgb}{0.65,0,0.65}
\definecolor{dark_green}{rgb}{0, 0.5, 0}
\definecolor{green}{rgb}{0, 1.0, 0}
\definecolor{orange}{rgb}{0.8, 0.6, 0.2}
\definecolor{red}{rgb}{0.8, 0.2, 0.2}
\definecolor{blueish}{rgb}{0.0, 0.7, 1}
\definecolor{light_gray}{rgb}{0.7, 0.7, .7}
\definecolor{pink}{rgb}{1, 0, 1}

\newcommand{\hide}[1]{{}} 

\newcommand{\Figure}[1]{Fig.~\ref{fig:#1}}

\newcommand{\Section}[1]{Sec.~\ref{sec:#1}}

\newcommand{\algoname}{HITNet}
\usepackage{currfile}

\newcommand{\CIRCLE}[1]{\raisebox{.5pt}{\footnotesize \textcircled{\raisebox{-.6pt}{#1}}}}
\usepackage{amsmath}

\usepackage{caption} 
\newcommand{\argmin}{\operatorname{argmin}}

\newcommand{\feature}{\mathcal{F}}
\newcommand{\featureWeight}{\boldsymbol{\theta}_{\feature}}

\newcommand{\decriptor}{\mathcal{D}}
\newcommand{\decriptorWeight}{\boldsymbol{\theta}_{\decriptor}}

\newcommand{\update}{\mathcal{U}}

\newcommand{\featureMaps}{\mathcal{E}}
\newcommand{\featureMap}{\mathbf{e}}
\newcommand{\tileMaps}{\mathcal{\tilde{E}}}
\newcommand{\tileMap}{\mathbf{\tilde{e}}}

\newcommand{\localMap}{\mathbf{d'}}

\usepackage[pagebackref=true,breaklinks=true,letterpaper=true,colorlinks,bookmarks=false]{hyperref}

\def\cvprPaperID{****} 

\begin{document}

\title{\algoname: Hierarchical Iterative Tile Refinement Network for Real-time Stereo Matching}

\author{Vladimir Tankovich, 
Christian H\"ane, 
Yinda Zhang,
Adarsh Kowdle,
Sean Fanello,
Sofien Bouaziz \\
{\small Google}\\
{\tt\small \{vtankovich, chaene, yindaz, adarshkowdle, seanfa, sofien\}@google.com \vspace{-0.75cm}}
}

\maketitle

\begin{strip}
\centering
    \includegraphics[width=\linewidth]{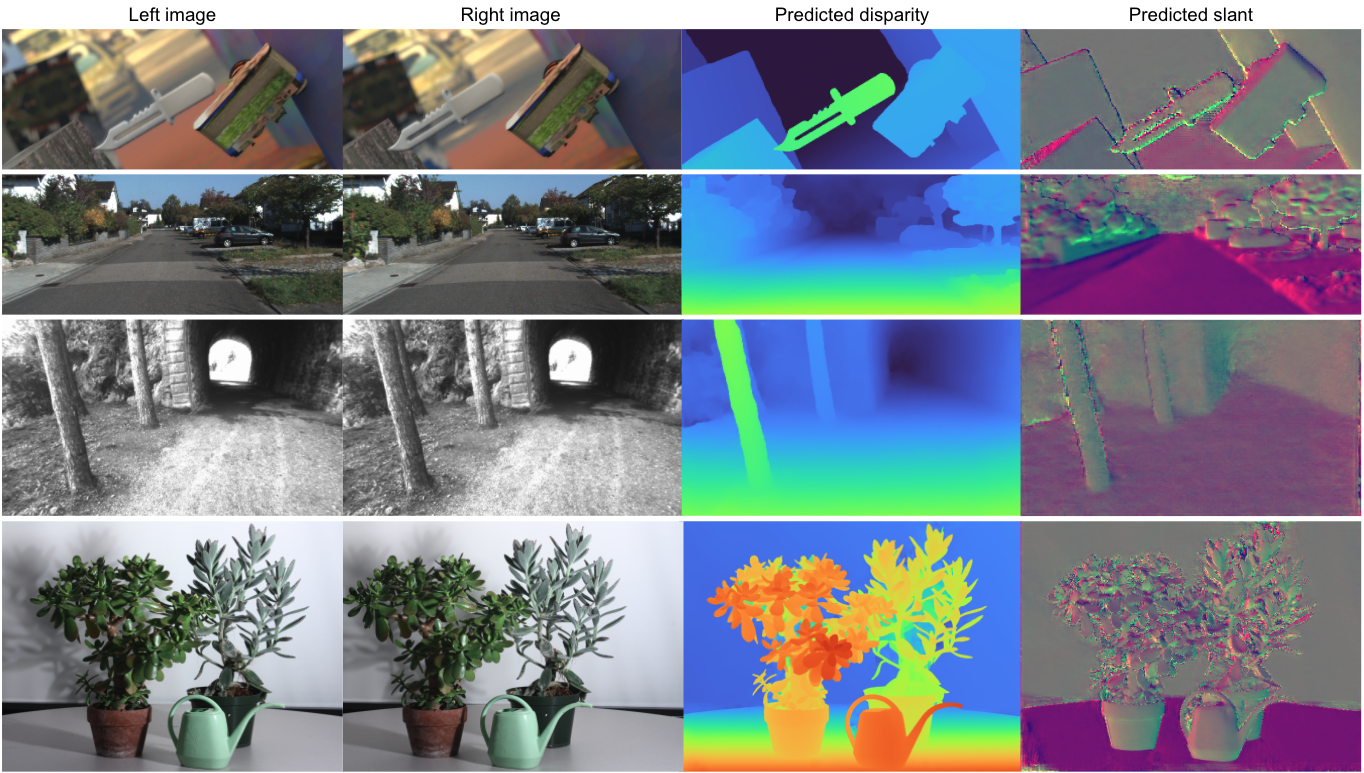}
\captionof{figure}{Example result of HITNet on the SceneFlow, KITTI, ETH3D and Middlebury datasets. Our approach predicts accurate depth with crisp edges. \algoname \ obtains state-of-art results on KITTI, ETH3D and Middlebury-v3 benchmarks.
\label{fig:teaser}}
\end{strip}

\begin{abstract}
This paper presents \algoname, a novel neural network architecture for real-time stereo matching. Contrary to many recent neural network approaches that operate on a full cost volume and rely on 3D convolutions, our approach does not explicitly build a volume and instead relies on a fast multi-resolution initialization step, differentiable 2D geometric propagation and warping mechanisms to infer disparity hypotheses. To achieve a high level of accuracy, our network not only geometrically reasons about disparities but also infers slanted plane hypotheses allowing to more accurately perform geometric warping and upsampling operations.
Our architecture is inherently multi-resolution allowing the propagation of information across different levels. Multiple experiments prove the effectiveness of the proposed approach at a fraction of the computation required by state-of-the-art methods. At the time of writing, \algoname \ ranks 1\textsuperscript{st}-3\textsuperscript{rd} on all the metrics published on the ETH3D website for two view stereo, ranks 1\textsuperscript{st} on most of the metrics amongst all the end-to-end learning approaches on Middlebury-v3, ranks 1\textsuperscript{st} on the popular KITTI 2012 and 2015 benchmarks among the published methods faster than $\mathit{100}$ms.
\end{abstract} 
\section{Introduction}
Recent research on depth from stereo matching has largely focused on developing accurate but computationally expensive deep learning approaches. Large convolutional neural networks (CNNs) can often use up to a second or even more to process an image pair and infer a disparity map. For active agents such as mobile robots or self driving cars such a high latency is undesirable and methods which are able to process an image pair in a matter of milliseconds are required instead. Despite this, only $4$ out of the top 100 methods on the KITTI 2012 leaderboard are published approaches that take less than $100$ms\footnote{Additional approaches faster than 100ms are on the leaderboard but the algorithms are unpublished and hence it is unknown how the results were achieved.}.

A common pattern in end-to-end learning based approaches to computational stereo is utilizing a CNN which is largely unaware of the geometric properties of the stereo matching problem. In fact, initial end-to-end networks were based on a generic U-Net architecture \cite{mayer2016large}. Subsequent works have pointed out that incorporating explicit matching cost volumes encoding the cost of assigning a disparity to a pixel, in conjunction with 3D convolutions provides a notable improvement in terms of accuracy but at the cost of significantly increasing the amount of computation \cite{kendall2017end}. Follow up work \cite{stereonet} showed that a downsampled cost volume could provide a reasonable trade-off between speed and accuracy. However, the downsampling of the cost volume comes at the price of sacrificing accuracy.

Multiple recent stereo matching methods \cite{sos,fanello17_hashmatch,need4speed} have increased the efficiency of disparity estimation for active stereo while maintaining a high level of accuracy. These methods are mainly built on three intuitions: Firstly, the use of compact/sparse features for fast high resolution matching cost computation; Secondly, very efficient disparity optimization schemes that do not rely on the full cost volume; Thirdly, iterative image warps using slanted planes to achieve high accuracy by minimizing image dissimilarity. All these design choices are used without explicitly operating on a full 3D cost volume. By doing so these approaches achieve very fast and accurate results for active stereo but they do not directly generalize to passive stereo due to the lack of using a powerful machine learning system. This therefore raises the question if such mechanisms can be integrated into neural network based stereo-matching systems to achieve efficient and accurate results opening up the possibility of using passive stereo based depth sensing in latency critical applications.

We propose \algoname, a framework for neural network based depth estimation which overcomes the computational disadvantages 
of operating on a 3D volume by integrating image warping, spatial propagation and a fast \textit{high resolution initialization step} into the network architecture, while keeping the flexibility of a learned representation by allowing features to flow through the network. The main idea of our approach is to represent image tiles as planar patches which have a learned compact feature descriptor attached to them. The basic principle of our approach is to fuse information from the high resolution initialization and the current hypotheses using spatial propagation. The propagation is implemented via a convolutional neural network module that updates the estimate of the planar patches and their attached features. In order for the network to iteratively increase the accuracy of the disparity predictions, we provide the network a local cost volume in a narrow band ($\pm 1$ disparity) around the planar patch using in-network image warping allowing the network to minimize image dissimilarity. To reconstruct fine details while also capturing large texture-less areas we start at low resolution and hierarchically upsample predictions to higher resolution. A critical feature of our architecture is that at each resolution, matches from the initialization module are provided to facilitate recovery of thin structures that cannot be represented at low resolution. An example output of our method shows how our network recovers very accurate boundaries, fine detail and thin structures in \Figure{teaser}.

To summarize, our main contributions are:
\begin{itemize}
    \item A fast multi-resolution initialization step that computes \textit{high resolution matches} using learned features.
    \item An efficient 2D disparity propagation that makes use of slanted support windows with learned descriptors.
    \item State-of-art results in popular benchmarks using a fraction of the computation compared to other methods.
\end{itemize}
\section{Related Work}

Stereo matching has been an active field of research for decades \cite{marr1976cooperative,scharstein2002taxonomy,hamzah2016literature}.
\textit{Traditional methods} utilize hand-crafted schemes to find local correspondences \cite{yoon2005locally,hosni2013fast,bleyer2008simple,hirschmuller2008stereo} and global optimization to exploit spatial context \cite{felzenszwalb2006efficient,klaus2006segment,kolmogorov2001computing}.
The run-time efficiency of most of these approaches are correlated with the size of the disparity space, which prevents real-time applications.
\textit{Efficient algorithms} \cite{li2015spm,lu2013patch,bleyer2011patchmatch,besse2014pmbp} avoid searching the full disparity space by using patchmatch \cite{barnes2009patchmatch} and {super-pixel~\cite{li2015spm}} techniques.
A family of machine learning based approaches, using random forest and decision trees, are able to establish correspondences quickly \cite{fanello14a,hyperdepth,fanello17_hashmatch,fanello2017ultrastereo}. However, these methods require either camera specific learning or post processing.
Recently, \textit{deep learning} brought big improvements to stereo matching. Early works trained siamese networks to extract patch-wise features or predict matching costs \cite{luo2016efficient,zbontar2016stereo,zagoruyko2015learning,zbontar2015computing}. 
End-to-end networks have been proposed to learn all steps jointly, yielding more accurate results \cite{shaked2017improved,mayer2016large,ranjan2017optical,song2020adastereo}. 

A key component in modern architectures is a cost volume layer \cite{kendall2017end} (or correlation layer \cite{ilg2017flownet}), allowing the network to run per-pixel feature matching.
To speed up computation, cascaded models \cite{pang2017cascade,chang2018pyramid,liang2017learning,gidaris2017detect,yang2019hierarchical,Zhang2019GANet} have been proposed to search in disparity space in a coarse-to-fine fashion. In particular, \cite{pang2017cascade} uses multiple residual blocks to improve the current disparity estimate. The recent work \cite{yang2019hierarchical} relies on a hierarchical cost volume, allowing the method to be trained on high resolution images and generate different resolutions \textit{on demand}. All these methods rely on expensive cost-volume filtering operations using 3D convolutions \cite{yang2019hierarchical} or multiple refinement layers \cite{pang2017cascade}, preventing real-time performance. Fast approaches \cite{stereonet,activestereonet} \textit{downsample} the cost volume in spatial and disparity space and attempt to recover fine details by edge-aware upsampling layers. These methods show real-time performance but sacrifice accuracy especially for thin structures and edges since they are missing in the low-res initialization.
 
Our method is inspired by classical stereo matching methods, which aim at propagating good sparse matches \cite{fanello17_hashmatch,fanello2017ultrastereo,sos}. In particular, Tankovich et al. \cite{sos} proposed a hierarchical algorithm that makes use of slanted support windows to amortize the matching cost computation in \textit{tiles}. Inspired by this work, we propose an end-to-end learning approach that overcomes the issues of hand-crafted algorithms, while maintaining computational efficiency. 

PWC-Net \cite{pwcnet}, although designed for optical flow estimation, is related to our approach.  The method uses a \textit{low resolution} cost volume with multiple refinement stages via image warps and local matching cost computations. Thereby following the classical pyramidal matching approach where a low resolution result gets hierarchically upsampled and refined by initializing the current level with the previous level's solution. In contrast we propose a fast, multi-scale, \textit{high resolution} initialization which is able to recover fine details that cannot be represented at low resolution. Finally, our refinement steps produce local slanted plane approximations, which are used to predict the final disparities, as opposed to standard bilinear warping and interpolation employed in \cite{pwcnet}.

\begin{figure*}[t]
    \centering
    \includegraphics[width=0.9\linewidth]{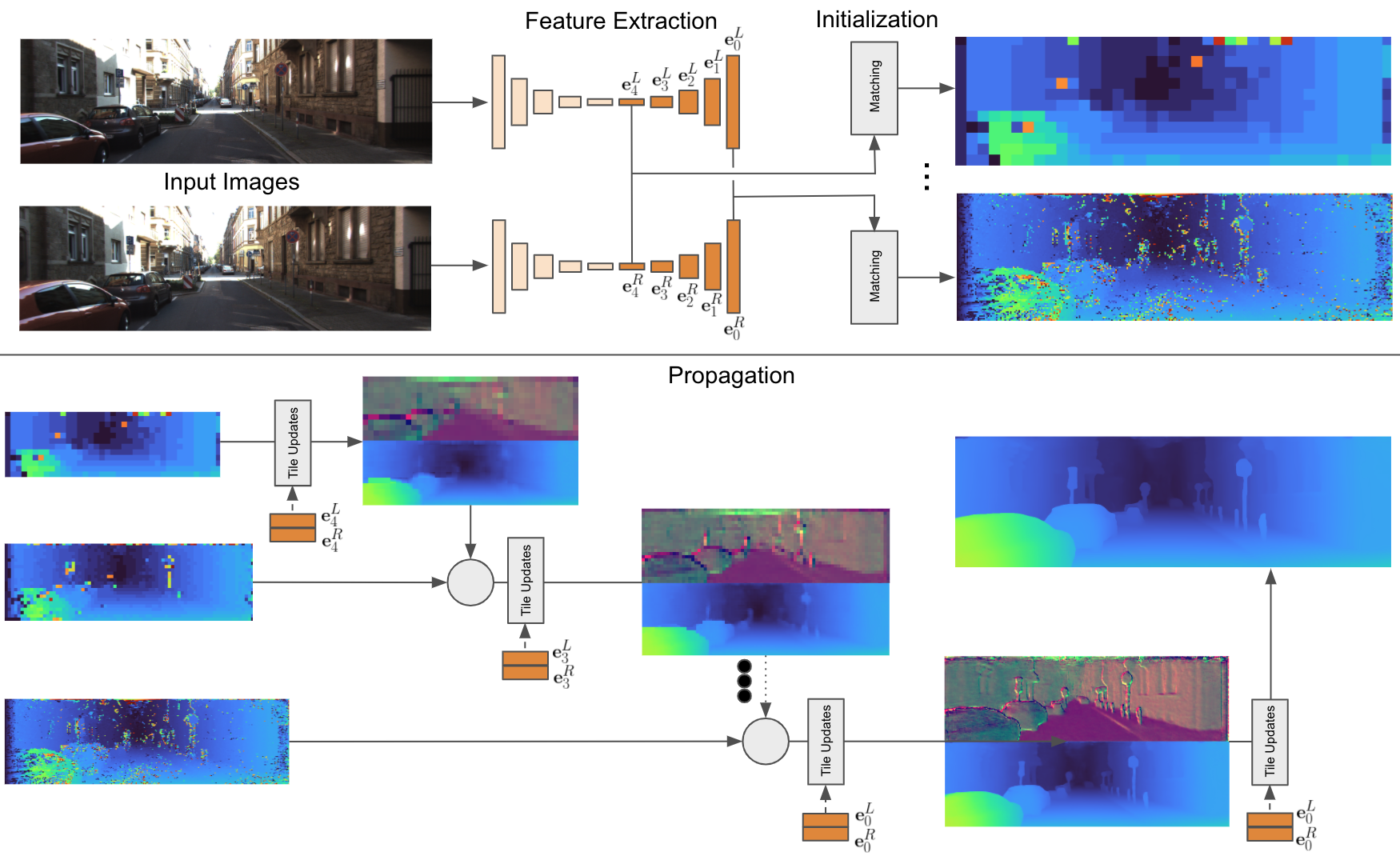}
    \caption{Overview of the proposed framework. (Top) A U-Net is used to extract features at multiple scales from left and right images. The initialization step is run on each scale of the extracted features. This step operates on \textit{tiles} of $4 \times 4$ feature regions and evaluates multiple disparity hypotheses. The disparity with the minimum cost is selected. (Bottom) The output of the initialization is then used at propagation stage to refine the predicted disparity hypotheses using slanted support windows.
   }
    \label{fig:overview}
    \vspace{-10pt}
\end{figure*}

\section{Method}
The design of \algoname, follows the principles of traditional stereo matching methods \cite{scharstein2002taxonomy}. In particular, we observe that recent efficient methods rely on the three following steps: \CIRCLE{1} compact feature representations are extracted \cite{fanello17_hashmatch,fanello2017ultrastereo}; \CIRCLE{2} a high resolution disparity initialization step utilizes these features to retrieve feasible hypotheses; \CIRCLE{3} an efficient propagation step refines the estimates using slanted support windows \cite{sos}. 
Motivated by these observations, we represent the disparity map as planar tiles at various resolutions and attach a learnable feature vector to each tile hypothesis (\Section{tile_hypothesis}). {This allows our network to learn which information about a small part of the disparity map is relevant to further improving the result.} This can be interpreted as an efficient and sparse version of the learnable 3D cost volumes that have shown to be beneficial \cite{kendall2017end}. 

The overall method is depicted in Fig.~\ref{fig:overview}. Our feature extraction module relies on a very small U-Net \cite{ronneberger2015u}, where the multi-resolution features of the decoder are used by the rest of the pipelines. These features encode multi-scale details of the image, similar to \cite{chang2018pyramid} (\Section{feature_extractor}).
Once the features are extracted, we initialize disparity maps as fronto parallel tiles at multiple resolutions. To do so, a matcher evaluates multiple hypotheses and selects the one with the lowest $\ell_1$ distance between left and right view feature. Additionally, a compact per-tile descriptor is computed using a small network (\Section{initialization}).
The output of the initialization is then passed to a propagation stage, which acts similarly to the approximated Conditional Random Field solution used in \cite{fanello17_hashmatch,sos}. This stage {hierarchically} refines the tile hypotheses in an iterative fashion (\Section{propagation}).

\subsection{Tile Hypothesis}
\label{sec:tile_hypothesis}
We define a tile hypothesis as a planar patch with a learnable feature attached to it. Concretely, it consists of a geometric part describing a slanted plane with the disparity $d$ and the gradient of disparity in $x$ and $y$ directions $(d_x, d_y)$, and a learnable part $\mathbf{p}$ which we call tile feature descriptor. The hypothesis is therefore described as a vector which encodes a slanted 3D plane,
\begin{equation}
    \mathbf{h} = [\underbrace{d, d_x, d_y}_{\mathrm{plane}}, \underbrace{\mathbf{p}}_{\mathrm{descriptor}}]
\end{equation}
The tile feature descriptor is a learned representation of the tile which allows the network to attach additional information to the tile. This could for example be matching quality or local surface properties such as how planar the geometry actually is. We do not constrain this information and learned it end-to-end from the data instead.

\subsection{Feature Extractor}
\label{sec:feature_extractor}
The feature extractor provides a set of multi-scale feature maps $\featureMaps = \{\featureMap_0, \hdots \featureMap_M\}$ that are used for initial matching and for warping in the propagation stage. We denote a feature map as $\featureMap_l$ and an embedding vector $\featureMap_{l,x,y}$ for locations $x,y$ at resolution $l \in {0,\ldots,M}$, $0$ being the original image resolution and $M$ denoting a  $2^M\times2^M$ downsampled resolution. A single embedding vector $\mathbf{e}_{l, x, y}$ is composed of multiple feature channels. We implement the feature extractor $\featureMaps = \feature (\mathbf{I};\featureWeight)$ as a U-Net like architecture \cite{ronneberger2015u,long2015fully}, i.e. an encoder-decoder with skip connections, with learnable parameters $\featureWeight$. The network is composed of strided convolutions and transposed convolutions with leaky ReLUs as non-linearities. The set of feature maps $\featureMaps$ that we use in the remainder of the network are the outputs of the upsampling part of the U-Net at all resolutions. This means that even the high resolution features do contain some amount of spatial context. In more details, one down-sampling block of the U-Net has a single $3\times3$ convolution followed by a $2\times2$ convolution with stride $2$. One up-sampling block applies $2\times2$ stride $2$ transpose convolutions to up-sample results of coarser U-Net resolution. Features are concatenated with skip-connection, and a $1\times1$ convolution followed by a $3\times3$ convolution are applied to merge the skipped and upsampled feature for the current resolution. Each upsampling block generates a feature map $\mathbf{e}_{l}$, which is then used for downstream tasks and also further upsampled in the U-Net to generate a higher resolution feature map.
We run the feature extractor on the left and the right image and obtain two multi-scale representations $\featureMaps^L$ and $\featureMaps^R$.

\subsection{Initialization}
\label{sec:initialization}
The goal of the initialization is to extract an initial disparity $d^{\mathrm{init}}$ and a feature vector $\mathbf{p}^{\mathrm{init}}$ for each tile at various resolutions. The output of the initialization is fronto-parallel tile hypotheses of the form $\mathbf{h}^{\mathrm{init}} = [d^{\mathrm{init}}, 0, 0, \mathbf{p}^{\mathrm{init}}].$

\paragraph{Tile Disparity.} 
In order to keep the initial disparity resolution high we use overlapping tiles along the $x$ direction, i.e. the width, in the right (secondary) image but we still use non-overlapping tiles in the left (reference) image for efficient matching.
To extract the tile features we run a $4 \times 4$ convolution on each extracted feature map $\mathbf{e}_l$. The strides for the left (reference) image and the right (secondary) image are different to facilitate the aforementioned overlapping tiles. For the left image we use strides of $4 \times 4$ and for the right image we use strides of $4 \times 1$, which is crucial to maintain the full disparity resolution to maximize accuracy. This convolution is followed by a leaky ReLU and an MLP.

The output of this step will be a new set of feature maps  $\tileMaps = \{\tileMap_{0}, \hdots, \tileMap_{M}\}$ with per tile features $\tileMap_{l,x,y}$. Note that the width of the feature maps in $\tileMaps^L$ and $\tileMaps^R$ are now different. The per-tile features are explicitly matched along the scan lines. We define the matching cost $\varrho$ at location $(x,y)$ and resolution $l$ with disparity $d$ as:
\begin{equation}
\varrho(l,x,y,d) =  \|\mathbf{\tilde{e}}_{l,x,y}^{L} - \mathbf{\tilde{e}}_{l,4x-d,y}^{R}\|_1
\end{equation}
The initial disparities are then computed as:
\begin{equation}
d^{\mathrm{init}}_{l,x,y} = \argmin_{d \in [0, D]} \varrho(l,x,y, d)
\end{equation}
for each $(x,y)$ location and resolution $l$, where $D$ is the maximal disparity that is considered. Note that despite the fact that the initialization stage exhaustively computes matches for all disparities there is no need to ever store the whole cost volume. At test time only the location of the best match needs to be extracted, which can be done very efficiently utilizing fast memory, e.g. shared memory on GPUs and a fused implementation in a single Op. Hence, there is no need to store and process a 3D  cost volume.

\paragraph{Tile Feature Descriptor.} The initialization stage also predicts a feature description $\mathbf{p}^{\mathrm{init}}_{l,x,y}$ for each $(x,y)$ location and resolution $l$:
\begin{equation}
    \mathbf{p}^{\mathrm{init}}_{l,x,y} = \decriptor(\varrho(d^{\mathrm{init}}_{l,x,y}), \tileMap^{L}_{l,x,y}; \boldsymbol{\theta}_{\decriptor_l}).
\end{equation}
The features are based on the embedding vector of the reference image $\tileMap^{L}_{l,x,y}$ and the costs $\varrho$ of the best matching disparity $d_{\mathrm{init}}$. We utilize a perceptron $\decriptor$, with learnable weights $\decriptorWeight$, which is implemented with a $1\times1$ convolution followed by a leaky ReLU. The input to the tile feature descriptor includes the matching costs $\varrho(\cdot)$, which allows the network to get a sense of the confidence of the match.

\subsection{Propagation}
\label{sec:propagation}

The propagation step takes tile hypotheses as input and outputs refined tile hypotheses based on spatial propagation of information and fusion of information. It internally warps the features from the feature extraction stage from the right image (secondary) to the left image (reference) in order to predict highly accurate offsets to the input tiles. An additional confidence is predicted which allows for effective fusion between hypotheses coming from earlier propagation layers and from the initialization stage. 

\paragraph{Warping.}
The warping step computes the matching costs between the feature maps $\featureMap^L_l$ and $\featureMap^R_l$ at the feature resolution $l$ associated to the tiles. This step is used to build a local cost volume around the current hypothesis. Each tile hypothesis is converted into a planar patch of size $4 \times 4$ that it originally covered in the feature map. We denote the corresponding $4 \times 4$ local disparity map as $\localMap$ with
\begin{equation}
\localMap_{i,j} = d + (i-1.5)d_x + (j-1.5)d_y,
 \label{eq:plane}
\end{equation}
for patch coordinates $i,j \in \{0,\cdots,3\}$. The local disparities are then used to warp the features $\featureMap^R_l$ from the right (secondary) image to the left (reference) image using linear interpolation along the scan lines. This results in a warped feature representation $\featureMap^{R'}_l$ which should be very similar to the corresponding features of the left (reference) image $\featureMap^L$ if the local disparity maps $\localMap$ are accurate. Comparing the features of the reference $(x,y)$ tile with the warped secondary tile we define the cost vector $\pmb{\phi}(\mathbf{e}, \mathbf{d'}) \in \mathbb{R}^{16}$ as:
\begin{equation}
 \pmb{\phi}(\featureMap_{l}, \mathbf{d'}) =  [c_{0,0}, c_{0,1}, \ldots, c_{0,3}, c_{1,0} \ldots c_{3,3}],
\end{equation}
where $
 c_{i,j} =  \|\mathbf{e}^L_{l, 4x + i, 4y + j} - \mathbf{e}^{R}_{l, 4x + i -\localMap_{i,j}, 4y + j} \|_1
$.

\paragraph{Tile Update Prediction.}
This step takes $n$ tile hypotheses as input and predicts deltas for the tile hypotheses plus a scalar value $w$ for each tile indicating how likely this tile is to be correct, i.e. a confidence measure. This mechanism is implemented as a CNN module $\update$, the convolutional architecture allows the network to see the tile hypotheses in a spatial neighborhood and hence is able to spatially propagate information. A key part of this step is that we augment the tile hypothesis with the matching costs $\pmb{\phi}$ from the warping step. By doing this for a small neighborhood in disparity space we build up a local cost volume which allows the network to refine the tile hypotheses effectively. Concretely, we displace all the disparities in a tile by a constant offset of one disparity $\mathbf{1}$ in the positive and negative directions and compute the cost three times. Using this let $\mathbf{a}$ be the augmented tile hypothesis map for input tile map $\mathbf{h}$:
\begin{equation}
    \mathbf{a}_{l,x,y} = [\mathbf{h}_{l,x,y},\underbrace{ \pmb{\phi}(\mathbf{e}_l, \mathbf{d'}-\mathbf{1}) , \pmb{\phi}(\mathbf{e}_l, \mathbf{d'}), \pmb{\phi}(\mathbf{e}_l, \mathbf{d'}+\mathbf{1})}_{\text{local cost volume}}],
\end{equation}
for a location $(x,y)$ and resolution $l$,  The CNN module $\update_l$ then predicts updates for each of the $n$ tile hypothesis maps and additionally $w^i \in \mathbb{R}$ which represent the confidence of the tile hypotheses:
\begin{equation}
 (\underbrace{\Delta \mathbf{h}^1_l, w^1, \ldots, \Delta \mathbf{h}^n_l, w^n}_{\text{hypotheses updates}}) = \update_l(\mathbf{a}^1_l,\ldots,\mathbf{a}^n_l; \boldsymbol{\theta}_{\update_l}).
\end{equation}
The architecture of $\update$ is implemented with residual blocks \cite{he2016deep} but without batch normalization. Following \cite{stereonet} we use dilated convolutions to increase the receptive field. Before running a sequence of residual blocks with varying dilation factors we run a $1\times1$ convolution followed by a leaky ReLU to decrease the number of feature channels. The update module is applied in a hierarchical iterative fashion (see \Figure{overview}). At the lowest resolution $l=M$ we only have $1$ tile hypothesis per location from the initialization stage, hence $n=1$. We apply the tile updates by summing the input tile hypotheses and the deltas and upsample the tiles by a factor of 2 in each direction. Thereby, the disparity $d$ is upsampled using the plane equation of the tile and the remaining parts of the tile hypothesis $d_x$, $d_y$ and $\mathbf{p}$ are upsampled using nearest neighbor sampling. At the next resolution $M-1$ we now have two hypotheses: the one from the initialization stage and the upsampled hypotheses from the lower resolution, hence $n=2$. We utilize the  $w^i$ to select the updated tile hypothesis with highest confidence for each location. We iterate this procedure until we reach the resolution $0$, which corresponds to tile size $4\times4$ and full disparity resolution in all our experiments. To further refine the disparity map we use the winning hypothesis for the $4\times4$ tiles and apply propagation module 3 times: for $4\times4$, $2\times2$, $1\times1$  resolutions, using $n=1$. The output at tile size $1 \times 1$ is our final prediction. More details about the network architecture are provided in the supplementary material.

\section{Loss Functions}
Our network is trained end-to-end with ground truth disparities $d^{\mathrm{gt}}$ utilizing the losses described in the remainder of this section. The final loss is a sum of the losses over all the scales and pixels: $\sum_{l,x,y} L_l^{\mathrm{init}} + L_l^{\mathrm{prop}} +  L_l^{\mathrm{slant}} + L_l^{\mathrm{w}}$.

\subsection{Initialization Loss}
Ground truth disparities are given with subpixel precision, however matching in initialization happens with integer disparities. Therefore we compute the matching cost for subpixel disparities using linear interpolation. The cost for subpixel disparities is then  given as
\begin{equation}
\psi(d) = (d-\lfloor d \rfloor)\varrho(\lfloor d \rfloor + 1) + (\lfloor d \rfloor + 1 - d) \varrho(\lfloor d \rfloor),
\end{equation}
where we dropped the $l,x,y$ subscripts for clarity. To compute them at multiple resolutions we maxpool the ground truth disparity maps to downsample them to the required resolution. 
We aim at training the features $\mathcal E$  to be such that the matching cost $\psi$ is smallest at the ground truth disparity and larger everywhere else. To achieve this, we impose an $\ell_1$ contrastive loss \cite{Hadsell2006}
\begin{equation}
    L^{\mathrm{init}}(d^{\mathrm{gt}}, d^{\mathrm{nm}}) = \psi(d^{\mathrm{gt}}) + \max(\beta - \psi(d^{\mathrm{nm}}), 0),
\end{equation}
where $\beta > 0$ is a margin, $d^{\mathrm{gt}}$ the ground truth disparity for a specific location and 
\begin{equation}
d^{\mathrm{nm}} = \argmin_{d \in [0, D]/{\{d: d \in [d^{\mathrm{gt}}-1.5, d^{\mathrm{gt}}+1.5]\}}} \varrho(d)
\end{equation}
the disparity of the lowest cost non match for the same location. This cost pushes the ground truth cost toward $0$ as well as the lowest cost non match toward a certain margin. In all our experiments we set the margin to $\beta = 1$. Similar contrastive losses have been used to learn the matching score in earlier deep learning based approaches to stereo matching \cite{zbontar2016stereo,luo2016efficient}. However, they either used a random non-matching location as negative sample or used all the non matching locations as negative samples, respectively.

\subsection{Propagation Loss}
During propagation we impose a loss on the tile geometry $d$, $d_x$, $d_y$ and the tile confidence $w$.  We use the ground truth disparity $d^{\mathrm{gt}}$ and ground truth disparity gradients $d^{\mathrm{gt}}_x$ and $d^{\mathrm{gt}}_y$, which we compute by robustly fitting a plane to $d^{\mathrm{gt}}$ in a $9 \times 9$ window centered at the pixel.
In order to apply the loss on the tile geometry we first expand the tiles to a full resolution disparities $\hat{d}$ using the plane equation $(d,d_x,d_y)$ analogously to Eq.~\ref{eq:plane}. The slant portion is also up-sampled to full resolution using nearest neighbor approach before slant loss is applied.
We use the general robust loss function $\rho(\cdot)$ from \cite{barron2019general} which resembles a smooth $\ell_1$ loss, i.e., Huber loss. Additionally, we apply a truncation to the loss with threshold $A$
\begin{equation}
L^{\mathrm{prop}}(d,dx,dy) = \rho(\min(|d^{\mathrm{diff}}|, A), \alpha, c),
\end{equation}
where $d^{\mathrm{diff}} = d^{\mathrm{gt}} - \hat{d}$. Further we impose a loss on the surface slant, as
\begin{equation}
L^{\mathrm{slant}}(d_x,d_y)  =  \left\| \begin{array}{c}d_x^\mathrm{gt}  -  d_x \\
d_y^\mathrm{gt}  -  d_y \end{array} \right\|_1\chi_{|d^{\mathrm{diff}}| < B},
\end{equation}
where $\chi$ is an indicator function which evaluates to $1$ when the condition is satisfied and 0 otherwise.
To supervise the confidence $w$ we impose a loss which increases the confidence if the predicted hypothesis is closer than a threshold $C_1$ from the ground truth and decrease the confidence if the predicted hypothesis is further than a threshold $C_2$ away from the ground truth.
\begin{equation}
    L^{\mathrm{w}}(w) = \max(1-w,0)\chi_{|d^{\mathrm{diff}}| < C_1} + \max(w,0)\chi_{|d^{\mathrm{diff}}| > C_2}
\end{equation}
For all our experiments $A = B = C_1 = 1$; $C_2 = 1.5$. For the last several levels, when only a single hypotheses is available, loss is applied to all pixels ( $A = \infty$).

\section{Experiments}
We evaluate the proposed approach on popular benchmarks showing competitive results at a fraction of the computational time compared to other methods. We consider the following datasets: SceneFlow \cite{mayer2016large},  KITTI 2012 \cite{Geiger2012CVPR}, KITTI 2015 \cite{Menze2015ISA}, ETH3D \cite{schoeps2017cvpr}, Middlebury dataset V3 \cite{middlebury14}. Following the standard evaluation settings we consider the two popular metrics: the End-Point-Error (EPE), which is the absolute distance in disparity space between the predicted output and the groundtruth; the $x$-pixels error, which is the percentage of pixels with disparity error greater than $x$. For the EPE computation on SceneFlow we adopt the same methodology of PSMNet \cite{chang2018pyramid}, which excludes all the pixel with ground truth disparity bigger than $192$ from the evaluation. Unless stated otherwise we use a HITNet with $5$ levels, i.e. $M=4$.

In this section we focus on comparisons with state-of-art on popular benchmarks, detailed ablation studies, run-time breakdown, cross-domain generalization and additional evaluations, are provided in the supplementary material. The trained models used for submission to benchmarks and evaluation scripts can be found at \href{https://github.com/google-research/google-research/tree/master/hitnet}{https://github.com/google-research/google-research/tree/master/hitnet}

\begin{figure*}[!htbp]
    \centering
    \includegraphics[width=0.97\linewidth]{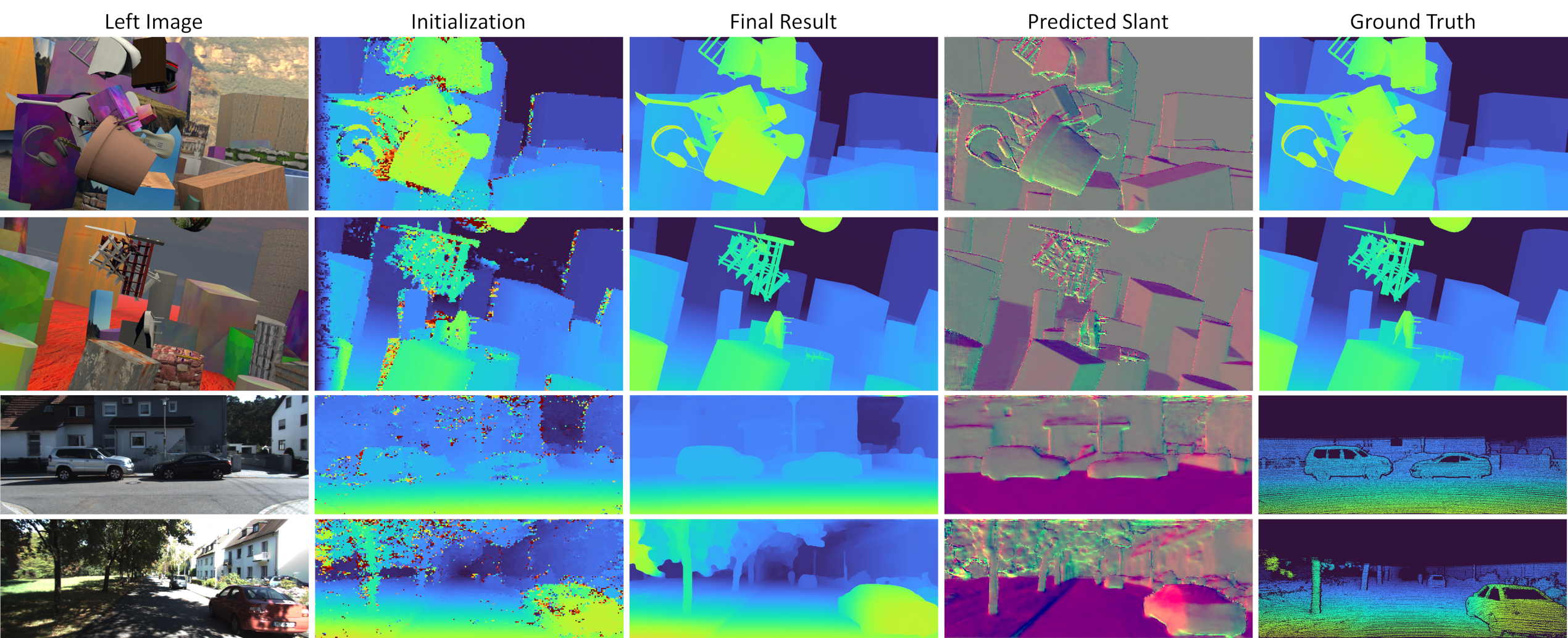}
    \caption{Qualitative results on SceneFlow and KITTI 2012. Note how the model is able to recover fine details, textureless regions and crisp edges. 
   }
    \label{fig:sceneflow}
    \vspace{-10pt}
\end{figure*}

\begin{table}[ht]
\begin{center}
\begin{tabular}{|c|c|c|}
\hline
 Method & EPE px & Runtime s \\
\hline
\algoname \ XL & 0.36 & 0.114 \\
\algoname \ L & 0.43 & 0.054 \\ 
EdgeStereo \cite{song2020edgestereo} & 0.74 & 0.32 \\
LEAStereo \cite{cheng2020hierarchical} & 0.78 & 0.3 \\
GA-Net \cite{Zhang2019GANet} & 0.84 & 1.6 \\
PSMNet \cite{chang2018pyramid} & 1.09 & 0.41 \\
StereoNet \cite{stereonet} & 1.1 & 0.015 \\ 
\hline
\end{tabular}
\caption{Comparisons with state-of-the-art methods on Scene Flow ``finalpass'' dataset, lower is better.}
\label{tab:sceneflow}
\end{center}
\vspace{-20pt}
\end{table}

\begin{table}[ht]
\begin{center}
\begin{tabular}{|c|c|c|c|c|}
\hline
 \small{Method} & \small{EPE px} & \small{Bad 1} & \small{Bad 2} & \small{Runtime}\\
\hline
    \small{ \algoname \ (ours)} & 0.20 & 2.79 & 0.80 & 0.02 s\\
    \small{R-Stereo} & 0.18 & 2.44 & 0.44 & 0.81 s\\ 
    \small{DN-CSS} & 0.22 & 2.69 & 0.77 & 0.31 s\\ 
    \small{AdaStereo} \cite{song2020adastereo} & 0.26 & 3.41 & 0.74 & 0.40 s\\
    \small{Deep-Pruner} \cite{duggal2019deeppruner} & 0.26 & 3.52 & 0.86 & 0.16 s\\
    \small{iResNet} \cite{liang2017learning} & 0.24 & 3.68 & 1.00 & 0.20 s\\
    \small{Stereo-DRNet} \cite{chabra2019stereodrnet} & 0.27 & 4.46 & 0.83 & 0.33 s\\
    \small{PSMNet} \cite{chang2018pyramid} & 0.33 & 5.02 & 1.09 & 0.54 s\\
\hline
\end{tabular}
\caption{Comparisons with state-of-the-art methods on ETH3D stereo dataset. For all metrics lower is better.}
\label{tab:eth3d}
\end{center}
\vspace{-15pt}
\end{table}

\begin{table*}
    \centering
    \begin{tabular}{|c|cccccc|ccc|c|}
    \hline
     & \multicolumn{6}{c|}{KITTI 2012 \cite{Geiger2012CVPR}} & \multicolumn{3}{c|}{ KITTI 2015 \cite{Menze2015ISA} } & \\
    \hline
    Method& 2-noc & 2-all & 3-noc & 3-all & \thead{EPE \\ noc} & \thead{EPE\\all} & D1-bg & D1-fg & D1-all & Run-time \\
    \hline
    \algoname \ (ours) & {2.00} & {2.65} &{1.41}  & {1.89} &  {0.4} & {0.5} & {1.74}  & {3.20} & {1.98} & 0.02s\\
    \hline
    LEAStereo \cite{cheng2020hierarchical} & 1.90 & 2.39 & 1.13 & 1.45 & 0.4 &0.5 & 1.40 & 2.91 & 1.65 & 0.3s \\
    GANet-deep \cite{Zhang2019GANet} & 1.89 & 2.50 & 1.19 & 1.6 & 0.4 &0.5 & 1.48 & 3.46 & 1.81 & 1.8s \\
    EdgeStereo-V2 \cite{song2020edgestereo} & 2.32 & 2.88 & 1.46 & 1.83 & 0.4 & 0.5  & 1.84 & 3.30 & 2.08 & 0.32s \\
    {GC-Net} \cite{kendall2017end} & {2.71} & {3.46} &1.77 & 2.30 & 0.6 & 0.7 & 2.21 & 6.16 & 2.87 & 0.9s \\
    {SGM-Net \cite{sgm-net}}  & 3.60 & 5.15 & 2.29 & 3.50 & 0.7  & 0.9 & 2.66 & 8.64 & 3.66  & 67s\\ 
    {ESMNet \cite{guo2019learning}}	& {3.65} &	{4.30} & {2.08} & {2.53} & {0.6} & {0.7} & 2.57 & 4.86 &	2.95 & 0.06s \\
    {MC-CNN-acrt \cite{zbontar2016stereo}}  & 3.90 & 5.45 & 2.09 & 3.22 & 0.6  & 0.7 & 2.89 & 8.88 & 3.89 & 67s\\
    RTSNet \cite{lee19} & 3.98 & 4.61 & 2.43 & 2.90 & 0.7 & 0.7 & 2.86 & 6.19 & 3.41 & 0.02s \\ 
    {Fast DS-CS \cite{yee2020fast}} & {4.54} & {5.34} & {2.61} & {3.20} & {0.7} &	{0.8} & 2.83 &	4.31 &	3.08 &	0.02s \\
    {StereoNet \cite{stereonet}} & 4.91 & 6.02 & - & - & 0.8 & 0.9 & 4.30 & 7.45 & 4.83 & 0.015s\\
    \hline
    \end{tabular}
    \caption{Quantitative evaluation on KITTI 2012 and KITTI 2015. For KITTI 2012 we report the percentage of pixels with error bigger than $x$ disparities in both non-occluded (x-noc) and all regions (x-all), as well as the overall EPE in both non occluded (EPE-noc) and all the pixels (EPE-all). For KITTI 2015 We report the percentage of pixels with error bigger than $1$ disparity in background regions (bg), foreground areas (fg), and all.}
\label{tab:kitti1215}
\vspace{-3pt}
\end{table*}

\begin{table*}
\begin{center}
\begin{tabular}{|c|c|c|c|c|c|c|c|c|}
\hline
 {Method} & {RMS} &  {AvgErr} & {Bad 0.5} & {Bad 1.0} & {Bad 2.0} & {Bad 4.0} & {A50} & {Run-time}\\
\hline
    { \algoname \ (ours)} & 9.97 & 1.71 & 34.2  & 13.3 & 6.46 & 3.81 & 0.40 & 0.14 s\\
    {LEAStereo \cite{cheng2020hierarchical}} & 8.11 & 1.43 & 49.5 & 20.8 & 7.15 & 2.75 & 0.53 & 2.9  s\\ 
    {NOSS-ROB} \cite{li2019} & 12.2 & 2.08 & 38.2 & 13.2 & 5.01 & 3.46 & 0.42 & 662s (CPU)\\
    {LocalExp} \cite{Taniai18} & 13.4 & 2.24 & 38.7 & 13.9 & 5.43 & 3.69 & 0.43 & 881s (CPU)\\
    {CRLE} \cite{CRLE} & 13.6 & 2.25 & 38.1 & 13.4 & 5.75 & 3.90 & 0.42 & 1589s (CPU)\\
    {HSM} \cite{yang2019hierarchical} & 10.3 & 2.07 & 50.7 & 24.6 & 10.2 & 4.83 & 0.56 & 0.51 s\\
    {MC-CNN} \cite{zbontar2016stereo} & 21.3 & 3.82 & 40.7 & 17.1 & 8.08 & 4.91 & 0.45 & 150 s\\
    {EdgeStereo} \cite{song2020edgestereo} & 9.84 & 2.67 & 55.6 & 32.4 & 18.7 & 10.8 & 0.72 & 0.35 s\\
\hline
\end{tabular}
\caption{Comparisons with state-of-the-art methods on Middlebury V3 dataset. For all metrics lower is better.}
\label{tab:middlebury}
\end{center}
\vspace{-20pt}
\end{table*}

\subsection{Comparisons with State-of-the-art}
\paragraph{SceneFlow.} On the synthetic dataset SceneFlow ``finalpass'' we achieve the remarkable End-Point-Error (EPE) of $0.36$, which is 2X better than state-of-art at time of writing (see supplementary materials for details of L and XL versions). Representative competitors are reported in Tab. \ref{tab:sceneflow}. The PSMNet algorithm \cite{chang2018pyramid} performs multi-scale feature extraction similarly to our method, but in contrast they use a more sophisticated pooling layer. Here we show that our architecture is more effective. Compared to GA-Net \cite{Zhang2019GANet}, we do not need complex message passing steps such as SGM. The results we obtain show that our strategy is also achieving a very similar inference. Finally, a representative fast method, StereoNet \cite{stereonet} is considered, which we consistently outperform.
As result our method achieves the lowest EPE while still maintaining real-time performance. See Figure \ref{fig:sceneflow} for qualitative results.

\paragraph{Middlebury Stereo Dataset v3.} We evaluated our method with multiple state-of-art approaches on the Middlebury stereo dataset v3, see Table \ref{tab:middlebury} and the official benchmark website.\footnote{See ``HITNet'' entry on the  \href{http://vision.middlebury.edu/stereo/eval3/}{official dataset website}.}. 
As we can observe \textit{we outperform all the other end-to-end learning based approaches on most of the metrics}, we rank among the top $10$ when considering also hand-crafted approaches and in particular we rank \textit{first} for bad $0.5$ and A50, \textit{second} for bad $1$ and \textit{avgerr}. In addition, we note that our average error is impacted by specifically one image, \textit{DjembL}, which is due to the fact that we do not explicitly handle harsh lighting variations between input pairs. For visual results on the Middlebury datasets and details regarding the training procedure we refer the reader to the supplementary material.

\paragraph{ETH3D two view stereo.} We evaluated our method with multiple state-of-art approaches on the ETH3D dataset, see Tab.~\ref{tab:eth3d}. At time of submission to benchmark, \algoname \ ranks 1\textsuperscript{st}-4\textsuperscript{rd} on all the metrics published on the website. In particular, our method ranks 1\textsuperscript{nd} on the following metrics: bad $0.5$, bad $4$, average error, rms error, $50\%$ quantile, $90\%$ quantile: this shows that \algoname \ is resilient to the particular measurement chosen, whereas competitive approaches exhibits substantial differences when different metrics are selected. See the submission website for details.\footnote{See the ETH3D Website at \href{https://www.eth3d.net/low_res_two_view}{https://www.eth3d.net/low\_res\_two\_view} for the complete metrics.}.

\paragraph{KITTI 2012 and 2015.} At time of writing, among the published methods faster than $100$ms, \algoname \ ranks {\#1} on KITTI 2012 and 2015 benchmarks. Compared to other state-of-the-art stereo matchers (see Tab.~\ref{tab:kitti1215}), our approach compares favorably to GC-Net \cite{kendall2017end}, \cite{pang2017cascade} and many others. Recent methods such as GA-Net \cite{Zhang2019GANet} and HSM  \cite{yang2019hierarchical} are obtaining slightly better metrics, although they require $1.8$ and $0.15$ seconds respectively. Note also that HSM  \cite{yang2019hierarchical} has been trained with additional external high resolution data. Similarly, GA-Net \cite{Zhang2019GANet} is pre-trained on SceneFlow and fine-tuned on KITTI benchmarks, whereas our approach is fully trained on the small data available on KITTI. Compared to fast methods such as StereoNet \cite{stereonet} and RTSNet \cite{lee19}, our method consistently outperforms them by a considerable margin, showing that it can be employed in latency critical scenarios without sacrificing accuracy.\footnote{See the KITTI Website at \href{http://www.cvlibs.net/datasets/kitti/eval_stereo.php}{http://www.cvlibs.net/datasets/kitti/eval\_stereo.php} for the complete metrics.}.

\section{Conclusion}
We presented \algoname, a real-time end-to-end architecture for accurate stereo matching. We presented a fast initialization step that is able to compute high resolution matches using learned features very efficiently. These tile initializations are then fused using propagation and fusion steps. The use of slanted support windows with learned descriptors provides additional accuracy.
We presented state-of-the art accuracy on multiple commonly used benchmarks.
A limitation of our algorithm is that it needs to be trained on a dataset with ground truth depth. To address this in the future we are planning to investigate self-supervised methods and self-distillation methods to further increase the accuracy and decrease the amount of training data that is required. A limitation of our experiments is that different datasets are trained on separately and use slightly different model architectures. To address this in the future, a single experiment is required that aligns with Robust Vision Challenge requirements.

\paragraph{Acknowledgements.} We would like to thank Shahram Izadi for support and enabling of this project.

{\small
\bibliographystyle{ieee_fullname}
\bibliography{references}
}

\clearpage

\appendix

\section{Training Details}
In this section we add additional details regarding the training procedure and discuss difference among datasets.

\subsection{Training Setup}
The SceneFlow dataset consists of 3 components (Flyingthings, Driving and Monkaa) and comes with a predefined train and test split with ground truth for all examples. Following the standard practice with this dataset we use the predefined train and test split for all experiments. We also used only FlyingThings part of the dataset, as Driving and Monkaa don't have corresponding TEST sets and including them into training hurts accuracy for both Sceneflow and when it's used to pre-train for Middlebury. When all 35k images are used to train a model, the PSM EPE of XL model is $0.41$ on "finalpass".
We considered random crops of $320 \times 960$ and a batch size of $8$, and a maximum disparity of $320$. We trained for $1.42$M iterations using the Adam optimizer, starting from a learning rate of $4e^{-4}$, dropping it to $1e^{-4}$, then to $4e^{-5}$, then to $1e^{-5}$ after $1$M, $1.3$M, $1.4$M iterations respectively. 
The general robust loss for SceneFlow experiments was applied with, $\alpha = 0.9, c = 0.1$. For all other experiments, $\alpha= 0.8, c = 0.5$.

For real world datasets such as KITTI 2012 and 2015 a training set with ground truth and a test set where the ground truth is not available is provided. For the benchmark submission we trained the network on all $394$ images available from both datasets. For ablation studies on the KITTI dataset we split training set into a train and validation set with 75\% of the data in the training set and 25\% of the data in the validation set.  We trained with data augmentation, batch-size of $4$ and random crops of $311 \times 1178$ and a maximum disparity of $256$. The training schedule followed the following step: $400$k iterations with learning rate $4e^{-4}$, followed by $8$k iterations with learning rate $1e^{-4}$, followed by $2$k iterations with learning rate $4e^{-5}$. Note that the network is not pre-trained on any other datasets as in \cite{yang2019hierarchical}, and a small training set is sufficient for our method to achieve good performance. 

\begin{figure*}[htb]
    \centering
    \includegraphics[width=0.95\linewidth]{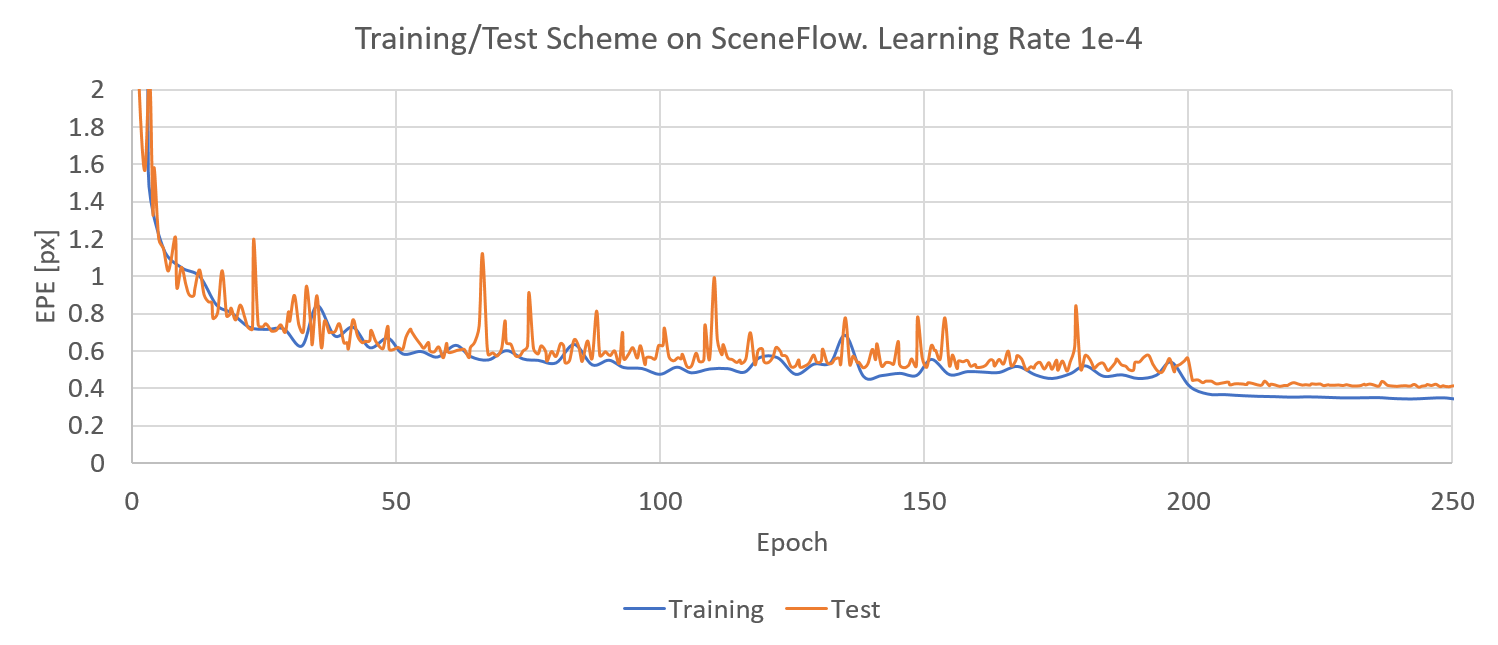}
    \caption{We show the evolution of the training reporting the EPE on training and test set respectively. Note how the scheme reduces the error on both training and test set without showing signs of overfitting.}
    \label{fig:training}
\end{figure*}

Indeed, empirically we found that using a small initial learning rate $1e^{-4}$ and training for longer achieves the best results on multiple datasets without showing sign of overfitting. In Figure \ref{fig:training} we show the evolution of the training \algoname \ L for more than $200$ epochs (learning rate change to $1e^{-5}$ after $200$ epochs) on the SceneFlow cleanpass dataset. We also compared this scheme with using a higher starting learning rate ($1e^{-3}$): after $10$ epochs we observed EPE of $0.85$ for $1e^{-4}$ and $0.66$ for $1e^{-3}$. Although $1e^{-3}$ achieved smaller error within a few epochs, our experiments confirm that longer training with a small learning rate is beneficial to achieve higher quality results without overfitting. See also generalization experiment showing that the method has very good cross-dataset performance.

The training set for the real world ETH3D stereo dataset \cite{schoeps2017cvpr} contains just a few stereo pairs, so additional data is needed to avoid overfitting. For the benchmark submission we trained the network on all $394$ images from both KITTI datasets, as well as all half and quarter resolution training images from Middlebury dataset V3 \cite{middlebury14} and training images from ETH3D dataset. We used the same training parameters as for KITTI submission and stopped training after $115$k iterations, which was picked using $4$ fold cross-validation on ETH3D training set. Note that there is no additional training, pre-training, finetuning.

Similarly, the Middlebury dataset \cite{middlebury14} contains a limited training set. To avoid overfitting, we pre-trained the model on SceneFlow's FlyingThings TRAIN set with data augmentation, then fine-tuned on the $23$ Middlebury14-perfectH training images, while keeping all data augmentations on. Specifically, we used a HITNet Large model with initialization at $6$ scales (M=$5$), pre-trained it for $445$k iterations, using batch size of $8$ and random crops of $512 \times 960$. 
We initialize the learning rate to $4e^{-4}$, then gradually drop it to $1e^{-4}$, $4e^{-5}$ and $1e^{-5}$ after $300$K, $400$K, $435$K iterations respectively. Finally, we fine-tuned the model for $5$K iterations at $1e^{-5}$ learning rate. These parameters were selected by using 4-fold cross validation on Middlebury training set.

\begin{figure}[t]
    \centering
    \includegraphics[width=\columnwidth]{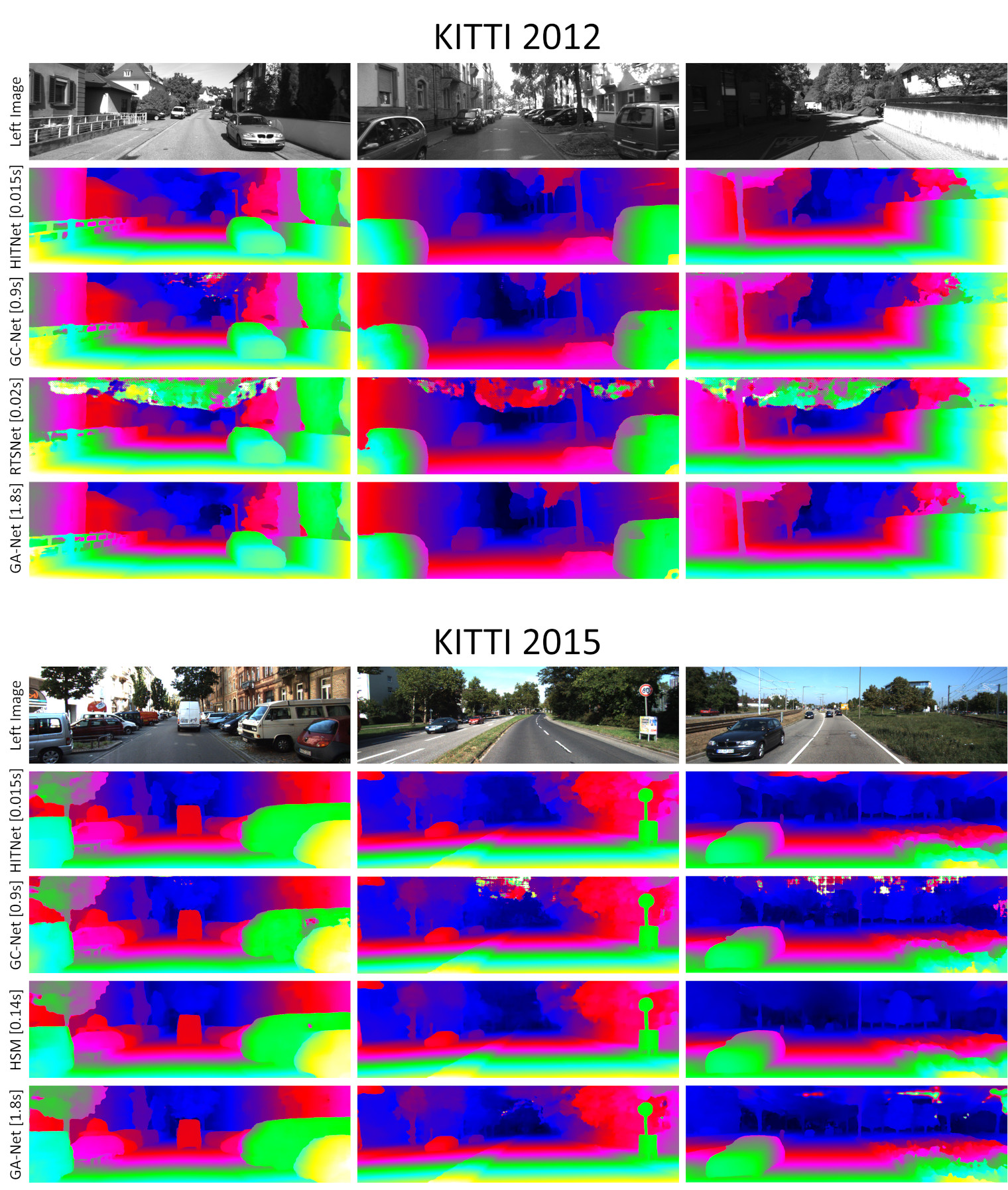}
    \caption{Qualitative Results on KITTI 2012 and 2015. Note how \algoname \ is able to recover fine structures and crisp edges using a fraction of the computational cost required by other competitors.}
    \label{fig:kitti}
\end{figure}

\subsection{Data Augmentation}
The training data available may not be fully representative of the actual test sets for small real world datasets such as KITTI, ETH3D and Middlebury. Indeed, we often observed substantial differences at test time, such as changes in brightness, unexpected reflections and mis-calibrations. In order to improve the network robustness we performed the following augmentations. We first perturb the brightness and contrast of left and right images by using random symmetric and asymmetric multiplicative adjustments. Symmetric adjustments are sampled within [0.8, 1.2] interval and asymmetric between [0.95, 1.05].  Similar to \cite{hsmnet}, We then replace random areas of the right image with random crops taken from another portion of the right image: this helps the network to deal with occluded areas and encourages a better ``inpainting''. The crop size to be replaced is randomly sampled between [50,50] and [180, 250].

Finally, the Middlebury images contains a substantially different color distribution compared to other datasets. To mitigate this we used the approach from \cite{song2020adastereo} that brings color distribution of training images closer to that of Middlebury set and during test time we normalize color distribution between left and right images of a stereopair. Additionally, similar to \cite{hsmnet}, in order to deal with miscalibrated pairs of this dataset, we augmented the training data with random y offset between $[-2,2]$ pixels. The random values for y offset are generated at a low resolution [H/64, W/64], and then bilineraly up-sampled to full resolution [H, W] of the input image. To simulate different noise levels images with different exposure contain, we add Gaussian random noise with variance sampled between [0 and 5] intensity levels once for the whole image.

\begin{figure*}[htb]
    \centering
    \includegraphics[width=0.95\linewidth]{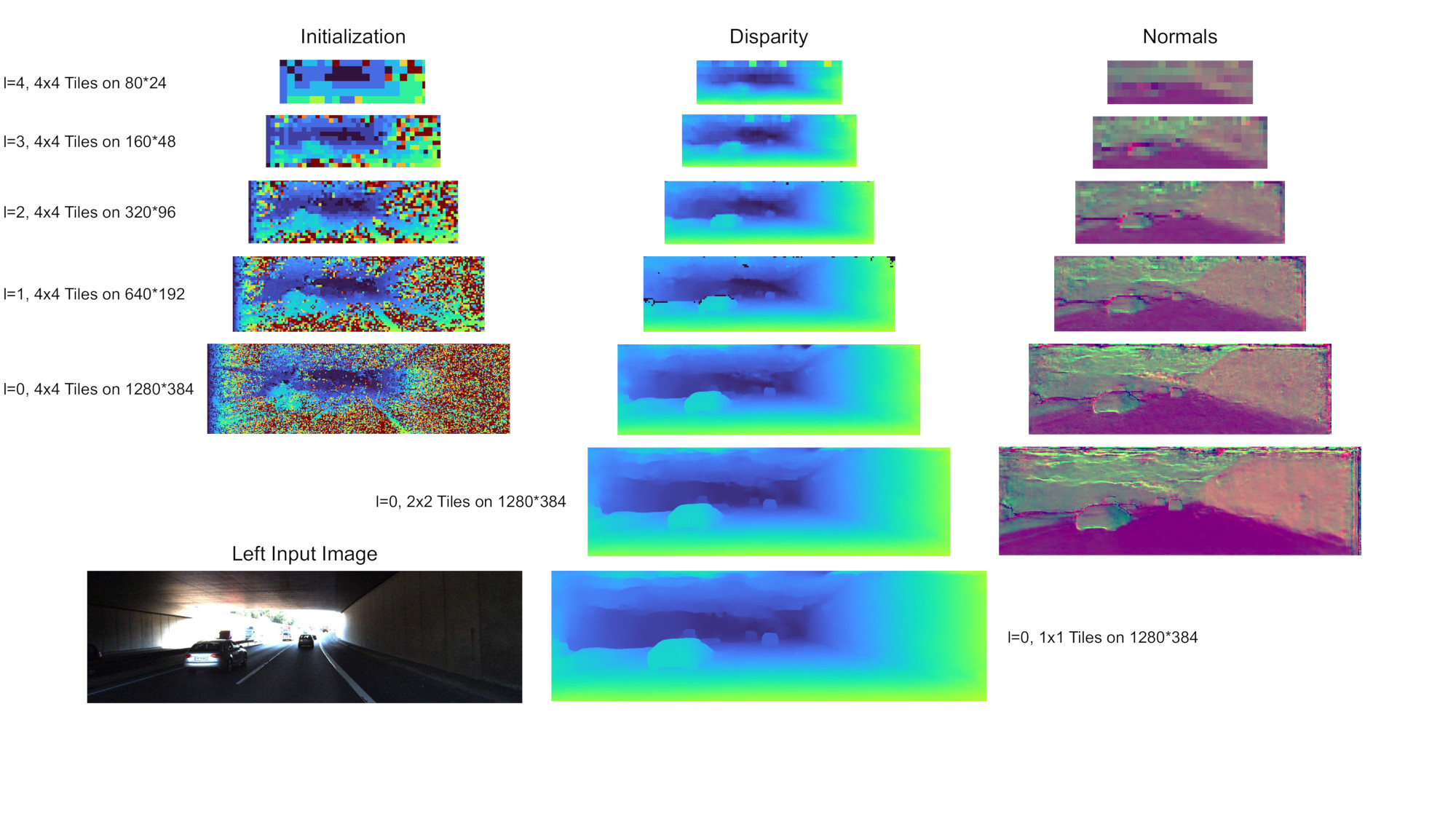}
    \vspace{-30pt}
    \caption{Intermediate results of our network on the left side we show the disparity maps that the matching of the initialization stage provides. On the right hand side we show the final disparity and normals for each resolution. The final two resolutions are 2x2 and 1x1 tiles of the highest resolution feature map, while the initialization is always computed on 4x4 tiles of the feature maps.}
    \label{fig:kitti_intermediate}
\end{figure*}

\begin{table*}[htb]
    \centering
    \begin{tabular}{|c|c|c|c|c|c|}
    \hline
     Dataset & \algoname  & CRL \cite{pang2017cascade} & iResNet \cite{Liang2018Learning} & PSMNet \cite{chang2018pyramid} & EdgeStereo \cite{song2020edgestereo} \\
    \hline
    KITTI 2012 EPE & {1.06}   & {1.38} &1.27 & 5.54 & 1.96\\
    KITTI 2012 $>3 px$  & 6.44  & 9.07& 7.89 & 27.33 & 12.27\\
    \hline
   KITTI 2015 EPE & {1.36} & {1.35} & 1.21 & 6.44 & 2.06\\
    KITTI 2015 $>3 px$ & 6.49  & 8.88& 7.42 &29.86 & 12.46\\
    \hline
    \end{tabular}
    \caption{Generalization Experiment. We trained each method on SceneFlow with data augmentation and tested on KITTI 2012 and 2015. Note how our method outperforms the others.}
\label{tab:generaliz}
\vspace{-10pt}
\end{table*}

\section{Additional Evaluations}

In this section we show additional qualitative results on real-world datasets. In Figure \ref{fig:kitti} we show comparisons of our method with other approaches. We consider multiple representative competitors such as: GC-Net \cite{kendall2017end}, which uses the full cost volume and 3D convolutions to infer context, RTS-Net \cite{lee19} that has similar inference time than HITNet, and finally GA-Net \cite{Zhang2019GANet}, as one of the best performing methods in terms of accuracy.

Our method compares very favorably to other approaches such as GC-Net and fast methods like RTSNet and is on par with the state-of-the-art approaches, e.g.\ GA-Net \cite{Zhang2019GANet}. Note how our method retrieves fine structures and crisp edges, while only training on the KITTI datasets, which exhibit significant edge fattening artifacts.

\begin{figure}[t]
    \centering
    \includegraphics[width=\columnwidth]{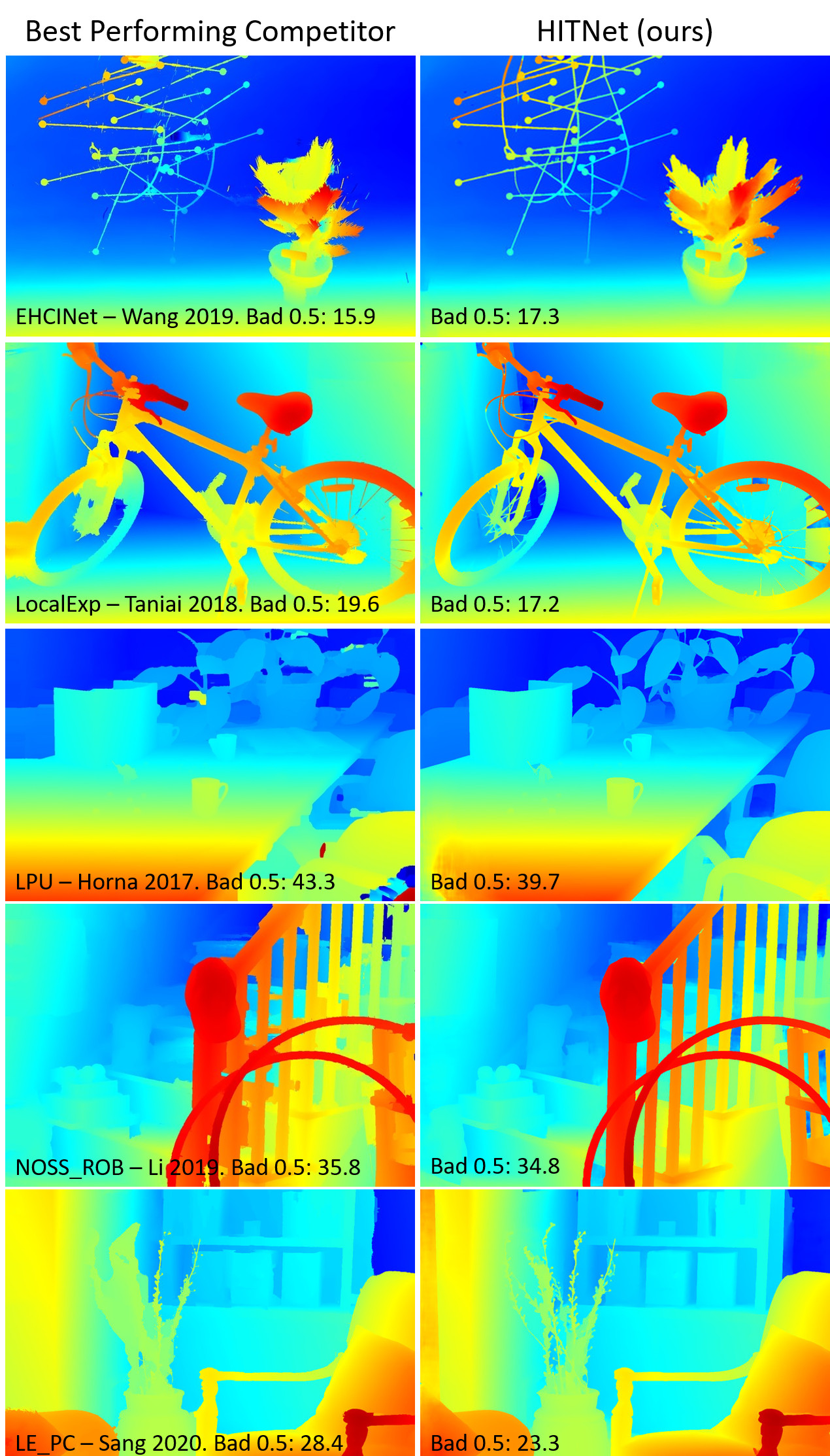}
    \caption{Qualitative comparisons on Middlebury dataset. For each image we compare our method with the best performing competitor following the Bad 0.5 metric. Note how our method is able to produce crisp edges, correct occlusions and thin structures in all the considered cases.}
    \label{fig:middlebury_qualitative}
    \vspace{-20pt}
\end{figure}

Similarly, in Figure \ref{fig:middlebury_qualitative} we show qualitative results on the Middlebury dataset \cite{middlebury14}. For each image, we compare \algoname \ with the best performing competitor on the Bad 0.5 metric. Note how our method is able to produce crisp edges, correct occlusions and thin structures in all the considered cases.

\begin{figure*}[htb]
    \centering
    \includegraphics[width=\linewidth]{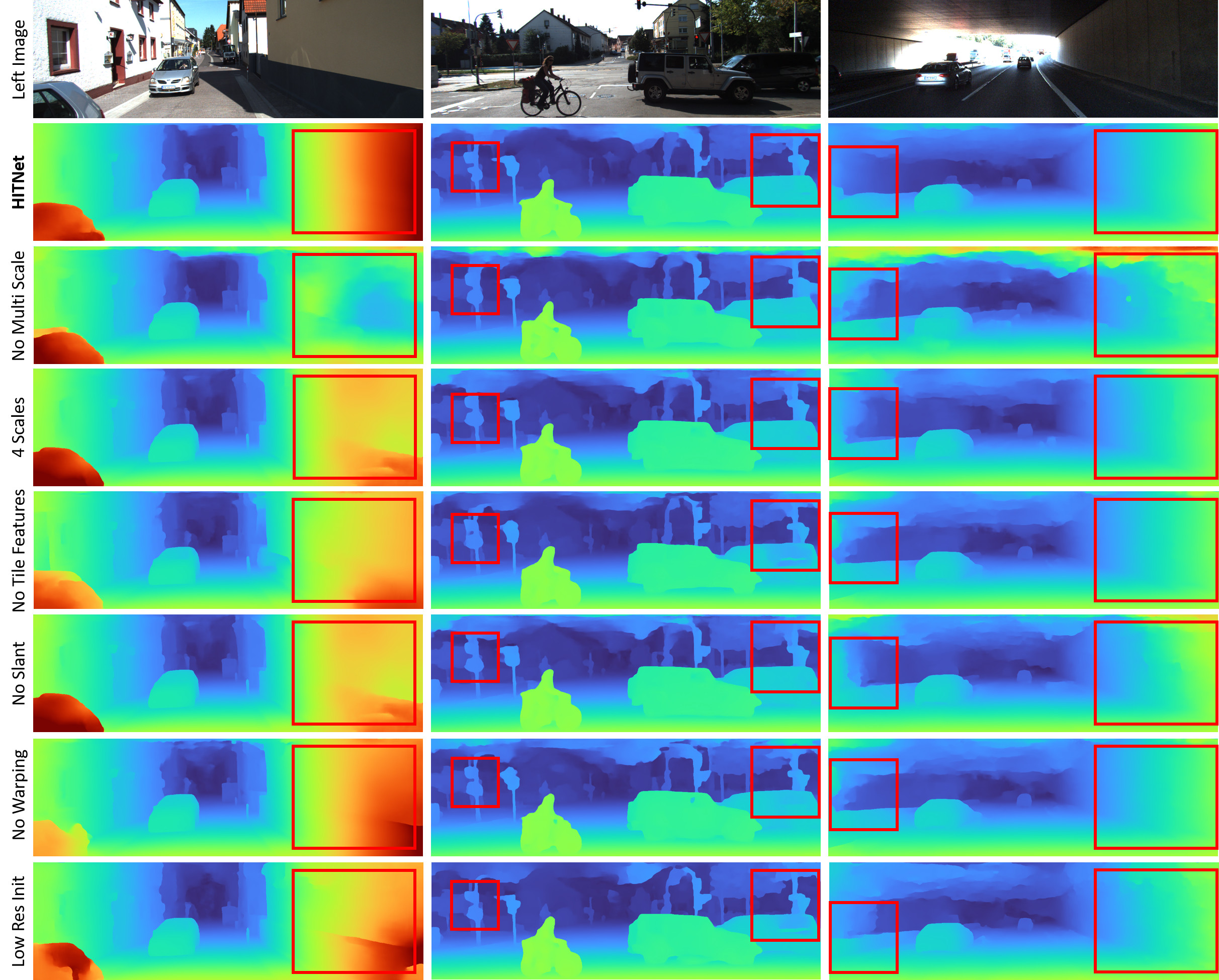}
    \caption{Ablation study, qualitative evaluation. Note how our \algoname \ model relies on all proposed design choices in order to achieve the best results on fine details, edges and occluded regions.
   }
    \label{fig:ablation}
\end{figure*}

\subsection{Intermediate Outputs}

We show intermediate outputs from within our network in Fig~\ref{fig:kitti_intermediate}. We observe that with increasing resolution the disparity gets more fine grained and the details from the higher resolution initialization gets merged into the global context that is coming from the lower resolutions. 
Note that our results on the KITTI 2015 dataset are only trained on the KITTI datasets from scratch without any pre-training on other data sources. This means the network has not been supervised on the top one third of the image as these datasets do only provide ground truth for the bottom two thirds of the image.

\subsection{Generalization.} We finally demonstrate the cross-domain adaptation capabilities of our method. Following the protocol in \cite{song2020edgestereo}, we trained \algoname \ on SceneFlow with data augmentations and tested on KITTI 2012 and KITTI 2015 respectively. We also considered multiple competitors as in \cite{song2020edgestereo} and report the results in Tab. \ref{tab:generaliz}: note how our method shows superior generalization results compared to all the other state-of-the-art approaches. This shows that our method is able to effectively generalize to unseen dataset even without explicit fine-tuning.

\begin{table*}[htb]
\begin{center}
\begin{tabular}{|c|c|c|c|c|c|c|c|}
\hline
&  \multicolumn{4}{c|}{ SceneFlow finalpass \cite{mayer2016large}} & \multicolumn{3}{c|}{ KITTI 2012 \cite{Geiger2012CVPR}} \\
\hline
Model & EPE & 0.1 px & 1 px & 3 px & EPE & 2 px & 3 px \\
\hline
\algoname \ & {0.529} px &  {24.0} \% & {5.52} \% &  {3.00} \%  &  {0.484 px}  &  {2.91} \% & {2.00} \% \\
$4$ Scales  & - & - & - & - &  {0.507} px & {3.10} \% &  {2.20} \%\\
No Multi-scale & - & - & - & - &  {0.747} px & {4.76} \% & {3.62} \%\\
$4$x$4$x$4$ Downsampled & {0.561} px &  {26.4} \% & {5.88} \%  &  {3.15} \% &  {0.526} px & {3.16} \% & {2.19} \%\\
$16$x$16$x$8$ Downsampled & {0.615} px &  {27.5} \% & {6.36} \%  &  {3.39} \% &  {0.536} px & {3.35} \% & {2.30} \%\\
$16$x$16$x$1$ Downsampled & {0.651} px &  {31.9} \% & {7.28} \%  &  {3.62} \% &  {0.554} px & {3.60} \% & {2.51} \%\\
No Warping &  {0.588} px &  {31.6} \% & {5.88} \%  &  {3.10} \%  &  {0.602} px &  {3.72} \% &  {2.54} \% \\
No Slant Prediction & {0.548} px &  {25.2} \% & {5.74} \%  &  {3.08} \%  & {0.513} px &  {3.23} \%  & {2.18} \%\\
No Tile Features &  {0.538} px &  {24.7} \% & {5.64} \%  &  {3.01} \%  &  {0.488} px &  {3.04} \%  &  {2.06} \%\\
\hline
\algoname \ L & {0.43} px & {20.7} \% & {4.70} \% & {2.57} \%  &  {0.490} px &  {2.98} \% &  {2.13} \% \\
\algoname \ XL & {0.36} px & {18.2} \% & {4.09} \% & {2.21} \%  &  {0.492} px &  {3.11} \% &  {2.20} \% \\
\hline
\end{tabular}
\end{center}
\caption{Ablation study of the proposed \algoname \ on SceneFlow \cite{mayer2016large} and KITTI 2012 \cite{Geiger2012CVPR} datasets. Lower is better.}
\label{tab:ablation}
\end{table*}

\subsection{Ablation Study}

We analyze the importance of our proposed components. The full \algoname \ is considered as baseline and compared with a version where features are removed. The ablation study is performed on the SceneFlow ``finalpass'' data and KITTI 2012. See Figure~\ref{fig:ablation} for a qualitative evaluation.

\paragraph{Multi Scale Prediction.} The multi-scale feature affects both initialization and propagation stages. In Tab. \ref{tab:ablation}, we report the results for the full model (\algoname) on KITTI 2012, with $5$ scales, results for $4$ scales and finally we removed the multi-resolution prediction completely. When we evaluated the same settings on the synthetic SceneFlow dataset we did not find a substantial differences between a single scale or multiple ones: clearly the synthetic dataset contains much more textured regions that do not benefit of additional context during propagation, whereas real world scenarios are full of textureless scenes (e.g. walls), where the multi-resolution approach is naturally performing better. 
\paragraph{$4$x$4$x$4$ Downsampled.} Initialization at full disparity resolution provides a compelling starting point to the network, which can focus mostly on refining the prediction. In Tab. \ref{tab:ablation} we show that using tile resolution for disparity (cost volume is 4X downsampled in H, W and D dimensions), the accuracy substantially drops. This demonstrates the importance of our proposed fast high resolution initialization.
\paragraph{$16$x$16$x$8$ Downsampled.} Decreasing the resolution of the cost volume for all dimensions similar to \cite{pwcnet} degrades accuracy (16X downsampled in H and W, 8X in D).
\paragraph{$16$x$16$x$1$ Downsampled.} Using larger tiles, while maintaining disparity resolution degrades accuracy even more, as the network is not able to reason about precise disparity at low spatial resolution during initialization.
\paragraph{Slant Prediction.} In this experiment, we forced tile hypotheses to always be fronto parallel by setting $d_x$ and $d_y$ to $0$ and using bilinear interpolation for upsampling. As showed in Tab. \ref{tab:ablation}, removing the slant prediction leads to a substantial drop in precision for both SceneFlow and KITTI 2012. Moreover the network loses its inherent capability of predicting some notion of surface normals that can be useful for many applications such as plane detection.
\paragraph{Tile Features.} Here we removed the additional features predicted on each tile during the initialization and propagation steps. This turns out to be a useful component and without it we observe a decrease in accuracy for both datasets.
\paragraph{Warping.} The image warps are used to compute the matching cost during the propagation. Removing this step hurts the subpixel precision as demonstrated in  Tab. \ref{tab:ablation}.
\paragraph{Model Size.} Finally, we tested if an increase in the model size is beneficial or not. In particular we double the channels in the feature extractor, and use 32 channels and 6 residual blocks for the last 3 propagation steps, this resorts to a run-time increase to 54ms. As expected this has an improvement on SceneFlow as reported in Tab. \ref{tab:ablation}, \algoname \ Large; however for the small KITTI datasets this did not improve performance due to over-fitting. Further increasing model size by using 64 channels for the last 3 propagation steps improved SceneFlow results, increased runtime to 114ms, and increased over-fitting on a smaller dataset. We don't see a reason to explore larger model sizes on a synthetic dataset as it will add to over-fitting on smaller real datasets that are publicly available. The metrics on "cleanpass" for XL version are: 0.31 epe, 15.6 bad 0.1, 3.67 bad 1.0, 1.99 bad 3.0.

\section{Model Architecture Details}
\label{sec:arch_details}
By default, the \algoname \ architecture is implemented with a 5-scale feature extractor with $16,16,24,24,32$ channels at corresponding resolutions. During Initialization step the first convolution over $4\times4$ tiles outputs 16 channels, followed by 2-layer MLP with 32 and 16 channels and ReLU non-linearities. Tile descriptor has $13$ channels by default, residual blocks use $32$ channels, unless mentioned otherwise. Each intermediate propagation steps use $2$ residual blocks without dilations. At each spatial resolutions, the propagation module uses feature maps from appropriate scale: full-resolution feature maps for $4\times4$ tiles, 2X downsampled for coarser tiles that have size $8\times8$ in full resolution, but sample $4\times4$ pixels in coarser feature map, etc till 16X downsampled and $64\times64$ tiles in original resolution. The last 3 levels of propagation start at $4\times4$ tiles and progressively in-paint and refine strong correct disparity at the edges over larger regions. To achieve that, they operate on coarse feature maps: the $4\times4$ tiles use 4X downsampled features for warping, the $2\times2$ tiles use 2X downsampled features for warping, the $1\times1$ tiles use full-resolution features for warping. 

In \algoname\ model used in KITTI and ETH3d experiments last 3 propagation steps use $4,4,2$ residual blocks with $32,32,16$ channels and $1,3,1,1$; $1,3,1,1$; $1,1$ dilations.

\algoname\ model used in Sceneflow experiments uses $16,16,24,24,32$ channels for feature extractor. A single initialization at 4x4 tiles. Last 3 propagation steps use $6,6,6$ residual blocks with $32,32,16$ channels and $1,2,4,8,1,1$ dilations.

\algoname L model used in Sceneflow experiments uses $32,40,48,56,64$ channels for feature extractor. A single initialization at 4x4 tiles. Last 3 propagation steps use $6,6,6$ residual blocks with $32,32,32$ channels and $1,2,4,8,1,1$ dilations.

\algoname XL model used in Sceneflow experiments uses $32,40,48,56,64$ channels for feature extractor. A single initialization at 4x4 tiles. Last 3 propagation steps use $6,6,6$ residual blocks with $64,64,64$ channels and $1,2,4,8,1,1$ dilations.

\algoname\ model used in Middlebury experiments uses $32,40,48,56,64$ channels for feature extractor. Last 3 propagation steps use $6,6,6$ residual blocks with $32,32,32$ channels and $1,2,4,8,1,1$ dilations.

The models used for submission for benchmarks and scripts to run them are available at \\ \href{https://github.com/google-research/google-research/tree/master/hitnet}{https://github.com/google-research/google-research/tree/master/hitnet}

\begin{figure}[h]
\vspace{-10pt}
    \centering
    \includegraphics[width=\linewidth]{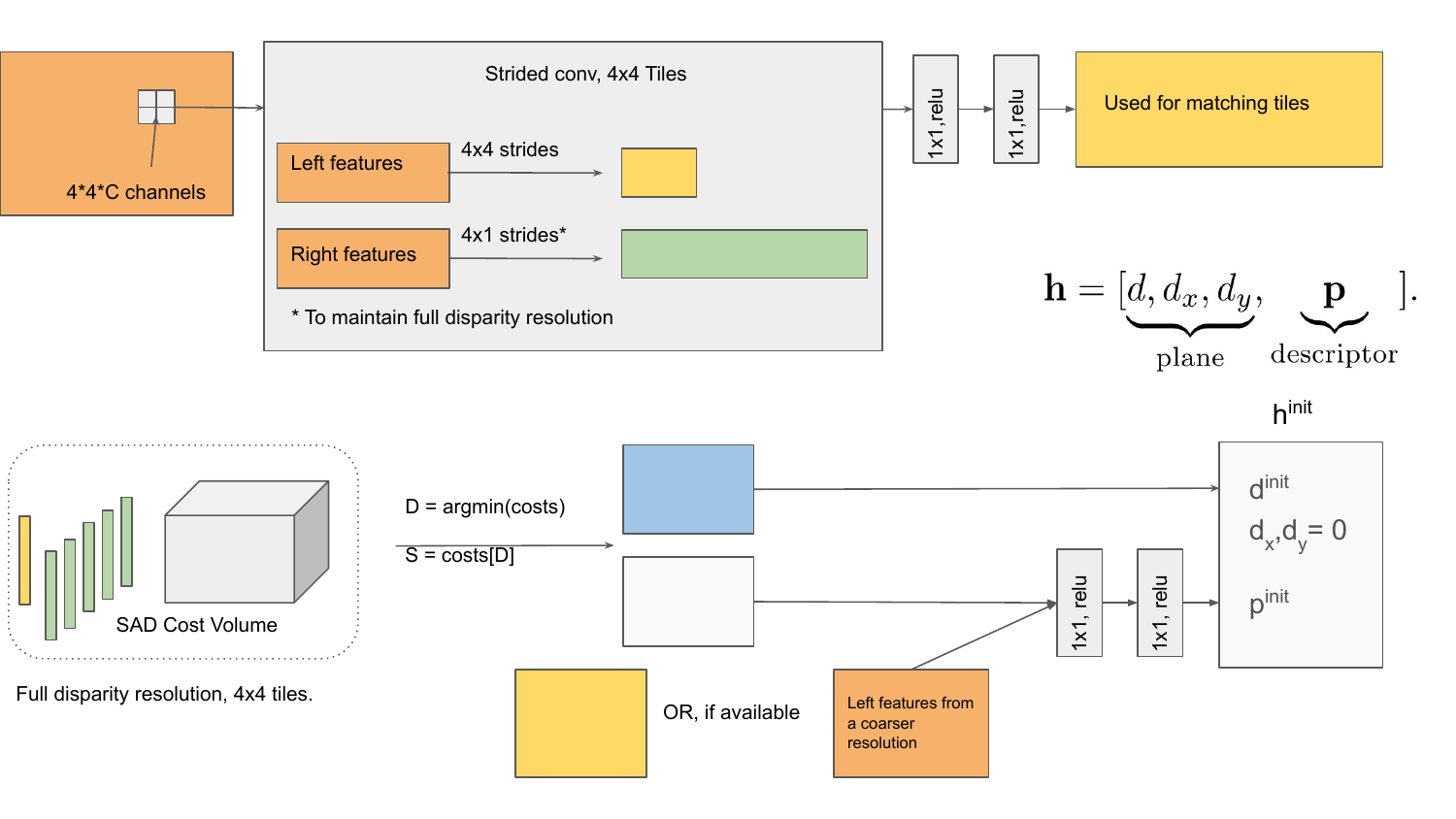}
    \caption{Initialization: The features extracted by the feature extractor are matched and inital tile features are computed. The number of feature channels $C$ depends on feature extractor architecture and the current level (see Sec.~\ref{sec:arch_details})}
    \label{fig:initialization}
\end{figure}

\begin{figure}[h]
\vspace{-10pt}
    \centering
    \includegraphics[width=\linewidth]{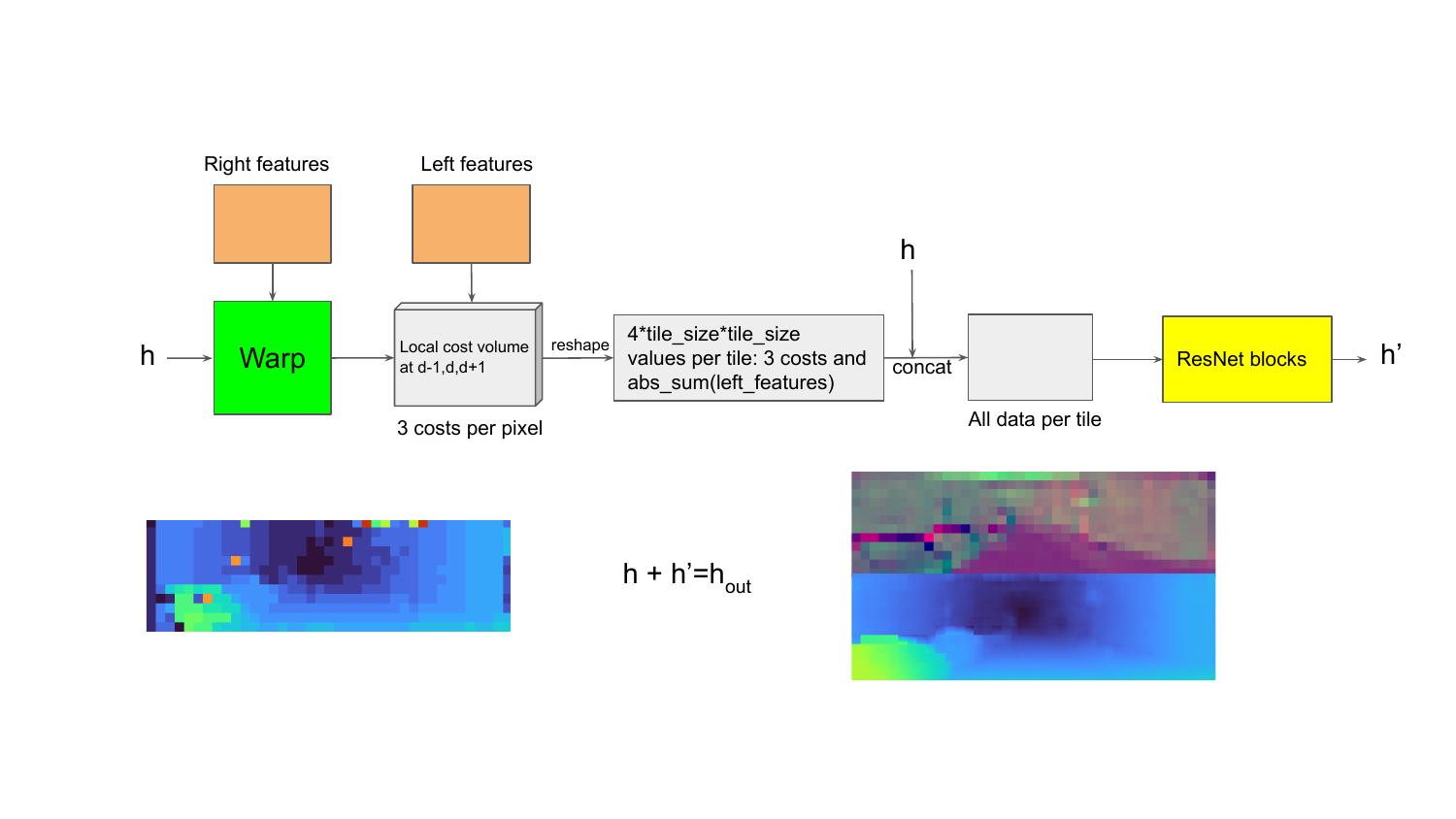}
    \caption{Propagation, for the single hypothesis case.}
    \label{fig:propagation_single}
\end{figure}

\begin{figure}[h]
\vspace{-10pt}
    \centering
    \includegraphics[width=\linewidth]{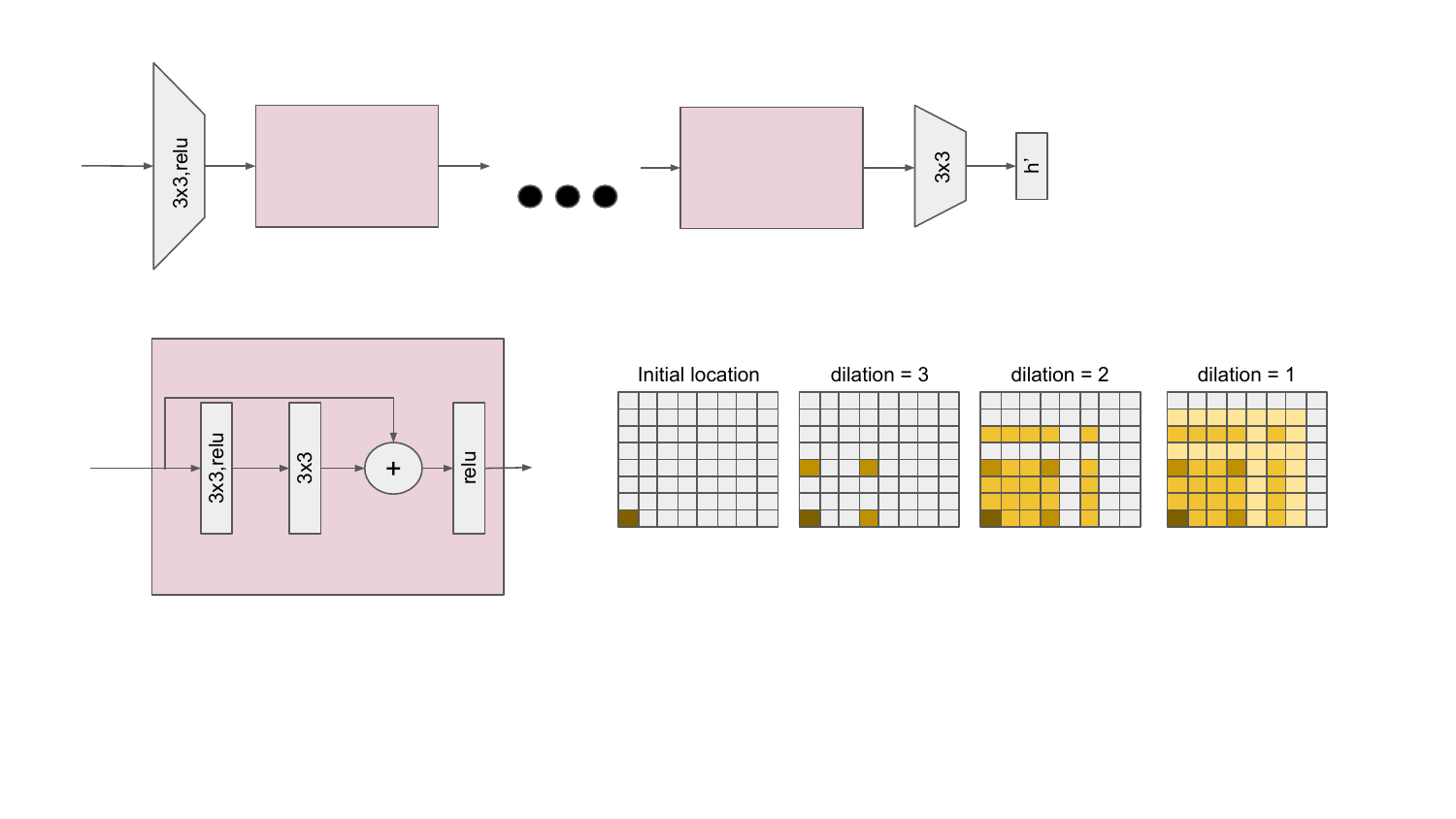}
    \caption{ResNet blocks. First 3x3 convolution generates a local state for the tile, which is incrementally updated by each block. The final state generates tile updates. Each block may use dilated convolutions to increase the speed of diffusion.}
    \label{fig:resblocks}
\end{figure}

\begin{figure}[h]
\vspace{-10pt}
    \centering
    \includegraphics[width=\linewidth]{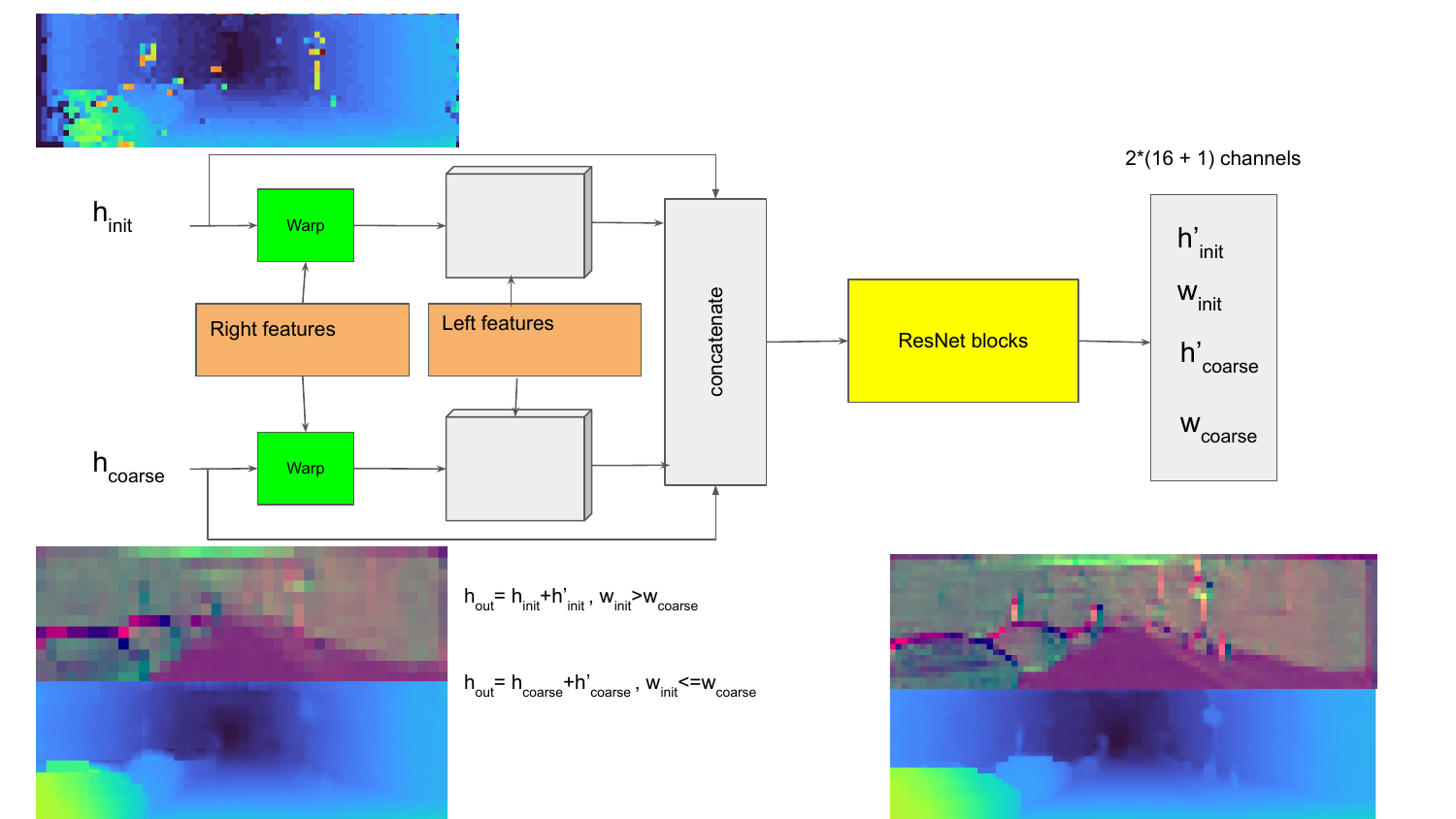}
    \caption{Propagation with multiple hypotheses.}
    \label{fig:propagation_multi}
\end{figure}

Fig~\ref{fig:initialization} provides more details for the initialization module described in the main paper. Similarly Fig~\ref{fig:propagation_single} and Fig~\ref{fig:resblocks} show how resblocks are integrated into propagation logic. Finally, Fig~\ref{fig:propagation_multi} depicts differences between propagation steps when a single hypothesis or multiple hypotheses are used.

\begin{table*}[htb]
\begin{center}
\begin{tabular}{|c|c|c|c|c|}
\hline
 Model & Param & GMac & EPE & 1 Pixel Threshold Error Rates \\
\hline
GC-Net \cite{kendall2017end} &  2.9M \cite{Zhang2019GANet} & 8789 \cite{chabra2019stereodrnet} & 1.80 \cite{Zhang2019GANet} & 15.6 \cite{Zhang2019GANet} \\
PSMNet \cite{chang2018pyramid} &  3.5M \cite{Zhang2019GANet} & 2594 \cite{chabra2019stereodrnet}& 1.09 \cite{Zhang2019GANet} & 12.1 \cite{Zhang2019GANet} \\
GANet \cite{Zhang2019GANet} & 2.3M \cite{Zhang2019GANet} & - & 0.84 \cite{Zhang2019GANet} & 9.9 \cite{Zhang2019GANet} \\
StereoDRNet \cite{chabra2019stereodrnet} & - & 1410 \cite{chabra2019stereodrnet} & 0.98 \cite{chabra2019stereodrnet} & - \\
LEAStereo \cite{cheng2020hierarchical} & 1.81M \cite{cheng2020hierarchical} & 782 \cite{cheng2020hierarchical} & 0.78 \cite{cheng2020hierarchical} & 7.82 \cite{cheng2020hierarchical} \\
\algoname \ Single-scale &  0.45M & 52 (92) & 0.53 & 5.52 \\
\algoname \ Multi-scale &  0.66M & 36 (61) & - & - \\
\algoname L &  0.97M & 146 (235) & 0.43 & 4.56 \\
\algoname \ Middlebury &  1.62M & 187; 450 for 1.57Mpix input & - & - \\
\algoname XL &  2.07M & 396 (735) & 0.36 & 4.09 \\
\hline
\end{tabular}
\caption{Comparisons of number of parameters and GMacs (Giga Multiply-accumulate operations) with other methods on Scene Flow ``finalpass'' dataset ($960\times540$ inputs). The numbers were partially adopted from the papers cited in the table. The lower the better. The multi-scale version of \algoname \ is used for ETH3d and KITTI submissions, GMac is provided for $1280\times384$ inputs. The GMac number in parenthesis is for predicting both disparity maps, sharing the feature extractor.}
\label{tab:parameters}
\end{center}
\vspace{-20pt}
\end{table*}

\section{Runing Time Details}
The \algoname \ architecture used for ETH3d and KITTI experiments runs at $19$ms per frame on a Titan V GPU for $0.5$Mpixel (KITTI resolution) input images. The majority of the time is spent during the last $3$ propagation steps ($7.5$ ms) that operate on higher resolutions. The multi-scale propagation steps use down-sampled data and contribute less than $5$ms. Efficient implementation of initialization using a single fused Op generates initial disparity estimates across all resolutions in $0.25$ms, with feature extractor contributing $6$ms.
For Middlebury experiments the model has a run-time of approximately $107.5$ms per Mpix of input resolution using custom CUDA operations for initialization and warping, when maximum disparity is $160$ and the run-time scales linearly with resolution. The run-time has a small increase with disparity range, and is about $109$ms per Mpix for a maximum disparity of $1024$. Without custom CUDA operations the run-time is increased by a factor of 3, as a single warping operation contains more than a hundred simple operations over large tensors, and while it's trivial to fuse them together, not doing so results in most of the time spent on global memory access. When tested on an 18-core Xeon 6154 CPU, the default tensorflow runtime runs $3.3$s per Mpix, which would translate to about $60s$ for a single threaded runtime, which compares favourably to other CPU methods. The CPU tensorflow runtime does make use of SIMD instruction set, which other methods may not utilize.

\section{Number of Parameters}

An important aspect of efficient neural network architectures is the number of parameters they have. This will influence the amount of compute required and the amount of memory needed to store them. Moreover, being able to achieve good performance with fewer numbers of parameters makes the network less susceptible to over-fitting. In Tab.~\ref{tab:parameters} we show that our network is able to achieve better results than other approaches with a significantly lower number of parameters and compute. 

Having less parameters also increases the generalization capabilities of the proposed method: indeed less learnable weights implies that the network is less prone to overfitting - our approach is able to outperform multiple state-of-the-art baselines when trained on synthetic data and tested in real-world scenarios.

\end{document}